\documentclass[10pt,twocolumn,letterpaper]{article}

\usepackage[pagenumbers]{wacv} 
\usepackage{graphicx}
\usepackage{amsmath}
\usepackage{amssymb}
\usepackage{booktabs}
\usepackage{float}
\usepackage{pifont}
\usepackage[pagebackref,breaklinks,colorlinks]{hyperref}

\usepackage[capitalize]{cleveref}
\crefname{section}{Sec.}{Secs.}
\Crefname{section}{Section}{Sections}
\Crefname{table}{Table}{Tables}
\crefname{table}{Tab.}{Tabs.}

\def\confName{WACV}
\def\confYear{2025}

\begin{document}

\title{GEM-VPC: A dual Graph-Enhanced Multimodal integration
for Video Paragraph Captioning}

\author{Eileen Wang$^1$ 
        \quad Soyeon Caren Han$^{1,2}$\thanks{\ \ Corresponding author. caren.han@sydney.edu.au} 
        \quad Josiah Poon$^1$\\ 
    $^1$School of Computer Science, The University of Sydney \\
    $^2$School of Computing and Information Systems, The University of Melbourne\\ 
    \texttt{\{Eileen.Wang, Caren.Han, Josiah.Poon\}@sydney.edu.au} \\
  }

\maketitle

\begin{abstract}
Video Paragraph Captioning (VPC) aims to generate paragraph captions that summarises key events within a video. Despite recent advancements, challenges persist, notably in effectively utilising multimodal signals inherent in videos and addressing the long-tail distribution of words. The paper introduces a novel multimodal integrated caption generation framework for VPC that leverages information from various modalities and external knowledge bases. Our framework constructs two graphs: a `\textit{video-specific}' temporal graph capturing major events and interactions between multimodal information and commonsense knowledge, and a `\textit{theme graph}' representing correlations between words of a specific theme. These graphs serve as input for a transformer network with a shared encoder-decoder architecture. We also introduce a node selection module to enhance decoding efficiency by selecting the most relevant nodes from the graphs. Our results demonstrate superior performance across benchmark datasets.
\end{abstract}

\section{Introduction}
Dense video captioning (DVC) \cite{krishna2017dense} is a sub-branch of video captioning, which requires the model to first localise the important events in the video and then generate the associated captions.
Video paragraph captioning (VPC) \cite{park2019adversarial} is a simplified version of DVC where the event segments in a video are assumed given; therefore, the event proposal generation step is not needed, and the ultimate goal is to generate better paragraph captions with the known events.
While research in video captioning is recently becoming more popular, numerous challenges still persist. Firstly, most VPC works solely use visual information for generating captions \cite{park2019adversarial, song2021towards}. However, they overlook that videos naturally contain rich content with multimodal signals such as additional speech text and an audio soundtrack. Incorporating these extra modalities and unravelling their interactions can provide vital cues for video understanding. Another challenge is overcoming the long-tail distribution of words, whereby the model tends to overfit on frequent terms while neglecting objects, properties or behaviours that rarely appear in the training data. Past natural language generation works have shown that exploiting external data from knowledge graphs can alleviate this issue and encourage more diverse generated text \cite{zhou2019improving}. 
Finally, existing studies \cite{iashin2020multi, lei2020mart} simply feed the video's feature embeddings into the captioning model directly, leading to two problems: 1) the model cannot effectively handle the long sequence, and 2) it struggles to select the relevant context from the long input stream.

As such, we address the aforementioned challenges by introducing GEM-VPC, a graph-based novel framework for VPC that integrates information from various modalities. Unlike past works \cite{iashin2020multi, iashin2020better}, rather than purely feeding in the raw features as a long input stream, we first convert the videos into a graphical structure to capture high-level salient features and context. We construct two types of graphs. The first is a `\textit{video-specific}' temporal graph, which aims to depict the major events of the video in chronological order whilst simultaneously representing interactions between various multimodal information and related commonsense knowledge. In particular, nodes are represented using language class labels to provide key details about the video contents instead of using raw feature embeddings, which may contain noisy information. To this end, we leverage pretrained action/audio/object recognition models and text parsers to extract linguistic information such as the action label, sound label or object label from the visual features, audio features and speech transcript to be used as nodes in the graph. To alleviate the long-tail problem, we further enhance the graph by incorporating language features from an external knowledge data source. While other VPC studies \cite{gu2023text} using knowledge graphs typically employ static graphs like ConceptNet \cite{speer2017conceptnet}, we use a neural knowledge model trained on existing commonsense knowledge graph datasets to generate diverse commonsense about human everyday experiences on-demand. These nodes are then connected with informative edge labels. We utilise sentences from the corpus to create a `\textit{theme graph}' to represent correlations between words relating to a specific theme with the motivation of providing corpus-level information for each sample during training. In the model training stage, both graphs are finally fed as supporting information into a transformer network. As some nodes in the graph may be noisy, we propose a node selection module to select only the most useful nodes from the video-specific and theme graphs when decoding the caption. 
\raggedbottom

The main contributions are to: 1) introduce a novel framework for VPC that leverages multimodal commonsense knowledge to enhance video understanding. It incorporates heterogeneous video and theme graphs derived from various modalities, including visual, audio, and textual data, along with commonsense knowledge. 2) demonstrate the superior performance of our model compared to state-of-the-art methods on two widely used benchmarks. 3) conduct a comprehensive ablation analysis to dissect the contribution of different components.

\section{Related Work\protect\footnote{T\lowercase{he main integration methods of past works are highlighted in} T\lowercase{able \ref{tab: automatic_metrics_anet_test} and} \ref{tab: automatic_metrics_yc2}}}

\subsection{Video Paragraph Captioning}

Earlier works for VPC often employ an LSTM-based model for generating the captions \cite{xiong2018move, zhang2018cross, zhou2019grounded}. \cite{park2019adversarial} adopts adversarial training in their LSTM model by proposing a hybrid discriminator to measure the language characteristics, relevance to a video segment, and coherence of their generated captions. Transformer-based \cite{vaswani2017attention} methods have become increasingly popular \cite{ging2020coot, wang2021end, yamazaki2023vltint, gu2023text}. This was first introduced by \cite{zhou2018end} for DVC and VPC, and each event in the video is decoded separately, resulting in context fragmentation and poor inter-event coherency. Later works have tried to alleviate this issue such as in MART \cite{lei2020mart}, which modified Transformer-XL \cite{dai2019transformer} and proposed a memory module for remembering the video segments and the sentence history to improve future caption predictions with respect to coherence and repetition aspects. \cite{yamazaki2023vltint} extracts local and global visual features and linguistic scene elements and leverages a Transformer to simultaneously model the long-range dependencies between features at an intra- and inter-event level.  

\subsection{Multimodal Video Captioning}
Existing studies have integrated multimodal features as extra information for video captioning. Most works consider the audio modality, with their frameworks first encoding the modalities separately with modality-specific encoders, followed by a fusion unit to combine the multiple streams together \cite{xu2017learning, rahman2019watch, iashin2020better}. Other than video and audio modalities, previous studies have suggested that considering speech features can enhance model outputs \cite{iashin2020multi}. In \cite{hessel2019case} and \cite{shi2019dense}, automatic speech recognition (ASR) was used to extract human speech from narrated instructional cooking videos for DVC while in \cite{gu2023text}, commonsense from knowledge graphs was incorporated into their captioning model where the ASR was used as a source for constructing the graph. Inspired by these methods, we consider the audio and speech modality as model inputs. Unlike the aforementioned approaches, we convert the videos into a heterogeneous graph from language labels extracted from the raw modality segments to represent relationships between key temporal events and different modality information, and propose a novel approach for explicitly incorporating the external commonsense knowledge into the graph. 


Some studies propose pretraining tasks to explicitly align the different modalities for improving feature representation, after which the model is fine-tuned to the captioning task. Common pretraining objectives involve predicting whether an ASR and video segment are aligned or predicting masked speech segments and frames \cite{huang2020multimodal, luo2020univl, li2020hero}. Generative pretraining objectives have been explored in \cite{yang2023vid2seq} and \cite{seo2022end}, which proposed predicting the transcribed speech given related video frames to jointly train the visual encoder and text decoder. Our framework requires no pretraining, but can achieve comparable scores to VPC models that utilise such methods.  

\subsection{Graphs for Video Analysis}
Graph structures have been widely used in video-related tasks from video scene graph classification \cite{arnab2021unified}, temporal action localisation \cite{zeng2019graph} to video question answering \cite{jiang2020reasoning} and visual storytelling \cite{wang-etal-2024-sco}. Several studies have delved into `spatio-temporal' graphs that try to represent interactions of features at a static time and relations between features across time. For the spatial component, numerous works connect objects and regions together within a timeframe and then connect identical or similar objects across time for the temporal component \cite{pan2020spatio, zhang2020does, jin2021adaptive, min2022learning}.  
In VPC, \cite{ji2022multimodal} proposed a multimodal heterogeneous graph that connects visual and text features within the same event. While they use the raw feature embeddings for node representation, which create large graphs with noisy information, we utilise the linguistic labels to provide a more high-level representation of the key semantic contents of the video and further propose a node selection module to filter out irrelevant nodes. 

\section{Method}
\textbf{Problem Definition}: Given an untrimmed video $v$ with temporally ordered events $E = \{e_{v1}, e_{v2},...,e_{vN}\}$ where $e_{vt}$ is the event at timestep $t$ defined by a starting and ending timestamp ($e_{vt}^s, e_{vt}^e$) and $N$ is the total number of events in the video, the task of VPC is to generate $Y = \{y_{v1}, y_{v2},...,y_{vN}\}$ where $y_{vt}$ is a matching textual description for $e_{vt}$.

We first describes constructing the graphs as input for our VPC model. Two graphs (Section \ref{vg_graph_creation} and \ref{tg_graph_creation}) are built: 1) a commonsense-enhanced video-specific graph (VG), representing the main sequential events in the video with related commonsense and contextual information, and 2) a theme graph (TG) representing relationships between vocabulary of a specific theme. For the video-specific graphs, we propose two ways to construct the primary nodes: 1) Utilising the video's visual information (`VF-method') and 2) extracting information from the speech transcript (`ASR-method').

\subsection{Video-Specific Graph Creation}\label{vg_graph_creation}
\subsubsection{Creating the Nodes - VF-Method}
Graphs created using the VF-method have 3 main node types: action, context (consisting of location, object, audio nodes), and commonsense nodes. 

\textbf{Action Nodes}: The action nodes describe the main actions at each key event and are represented using linguistic action class labels. To obtain these labels, we download the video frames at 5fps. For each event $e_{vt}$, we uniformly sample frames between the event's starting and ending frames with a step size of 10 and then feed every 16 frames into a pretrained video action classification model for each 16-frame segment. As the agent does not always perform a specific action (e.g. just standing or no human agent in the video segment), we replace the class label with `\textit{no action}' if the predicted class probability is less than a threshold. When less than the threshold and speech is detected by the audio node, we replace the label with `\textit{speaking}'. 

\textbf{Context Nodes}: For extra scene context, we include location, object and audio nodes. For the location and object nodes, we take the centre and last frame of each event and leverage a Visual Question Answering (VQA) model to extract open-ended answers about the images. For the location node, we ask the VQA model \textit{`what is the location?'} for each of the 3 images and take the most common answer as the location for each event. For the object nodes, we obtain the object labels by asking 3 questions: \textit{`what objects are in this image?'}, \textit{what is in the background?'} and \textit{`who is in this image?'}. We further expand this object set by employing an object detection model to detect objects from the frames. Finally, the audio nodes represent the sound information and can provide vital cues for video understanding in addition to the visual information. We sample 10 second segments of audio data from the video and obtain the the top 2 predicted audio classes by confidence score for each segment via a pretrained audio classifier. 

\textbf{Commonsense Nodes} We also add external commonsense knowledge for richer graphs. Comet-ATOMIC2020 \cite{hwang2021comet}, a \textit{neural knowledge model} capable of dynamically generating commonsense about everyday events is adopted. Given a head phrase and relation (e.g. cut a cake \texttt{CapableOf}), Comet-ATOMIC2020 can produce a tail phrase on-demand (e.g. celebrate birthday). We use the action node class labels as the head phrase and append 11 different relation tokens to generate 5 commonsense inferences per relation. The relation is described in Appendix E.
 
\subsubsection{Creating the Nodes - ASR-Method}
For videos where the speech modality is considered vital for video understanding, we introduce the ASR-method for creating the VG nodes. This is useful for how-to or cooking videos, where actions are explicitly described in the speech transcript, and visual information such as the location/scene may not be as important. There are 3 node types:

\textbf{Action Nodes:} We extract the ASR between each event and use a pretrained Open Information Extraction (OpenIE) model to breakdown the syntactically complex speech sentences into a list of verbs (V) and related arguments (ARG). Given the sentence \textit{`I chop the onions and put the meat in the frying pan'}, OpenIE can extract related arguments for the 2 verbs (\textit{`chop'} and \textit{`put'}): $<$ARG0, V, ARG1$>$ = $<$I, chop, onions$>$ and $<$ARG0, V, ARG1, ARG2$>$ = $<$I, put, meat, in the frying pan$>$. The extracted verb and argument tuples from the speech segments within each event are then used as the action nodes for event $e_i$. As the speech may contain irrelevant content, we tag the verbs in the ground-truth annotations and only retain tuples if the extracted verb has a high word embedding similarity score with at least one of the tagged verbs in the annotations. Moreover, we only retain words from the extracted arguments if it is a noun/adverb in the training annotations.


\textbf{Context Nodes:} Instead of location nodes as introduced in the VF-method, we concatenate the action node labels within the same event to form a `contextual phrase node'. This represents similar information to the action nodes, but at a less fine-grained level with more context about surrounding actions. For the object nodes, we tag the nouns from the ASR segment, retaining only the tagged nouns if they appear in the training ground-truth annotations. The audio nodes are retrieved in the same way as the VF-method except we filter out any irrelevant sound labels. For example, with cooking videos, we retain cooking-related sounds (`\textit{boiling}', `\textit{sizzling}', `\textit{frying}', `\textit{chopping}' etc). 

\textbf{Commonsense Nodes} We follow the VF-method but instead of using the action node information as the head phrase, we find that better commonsense is generated when using the linguistic information inside the contextual phrase node to query Comet-ATOMIC2020.

\subsubsection{Connecting the VG Nodes}
For event $e_{vt}$, let $AC_{t} = \{ac_{t1},...,ac_{tk}\}$ be the action nodes, $l_t$ be the corresponding location node when the VF-method is used,  or $cp_t$ be the contextual phrase node when the ASR-method is used, $CK_t=\{ck_{t1},...,ck_{tm}\}$ are the commonsense nodes, $O_t=\{o_{t1},...,o_{tn}\}$ are the object nodes, and $AU_t=\{au_{t1},...,au_{tp}\}$ are the audio nodes. 

To form the graph, all action nodes are first connected in temporal order. To capture forward information, we add a directed edge with the label \texttt{occursAfter} between each consecutive action node and further capture backwards information by adding a reversed edge with the label \texttt{occursBefore}. Each location node $l_t$ or contextual phrase node $cp_t$ is then connected to all the nodes in $AC_t$ with the edge label \texttt{atLocation} or \texttt{hasContext}. Next, commonsense nodes from $CK_t$ are connected to the corresponding action nodes from $AC_t$ that were used to generate the commonsense, using the commonsense relation token as the edge label. For the object and audio nodes, each node in $O_t$ and $AU_t$ is connected with $l_t$ or $cp_t$ with the edge label \texttt{inScene} and \texttt{hasSound} respectively. For the VF-method, we additionally filter out any irrelevant commonsense if the predicted action class confidence score used to generate that commonsense does not exceed a particular threshold. Noisy audio or object labels are disregarded at each timestep by converting the class labels to a text embedding and only retaining those that have a high cosine similarity score with any of the nodes in $AC_t$, $CK_t$ or $l_t$. A depiction of the final graphs using the VF- and ASR-method is in Appendix I.

\begin{figure*}[t]
  \centering
  \includegraphics[width=1\linewidth]{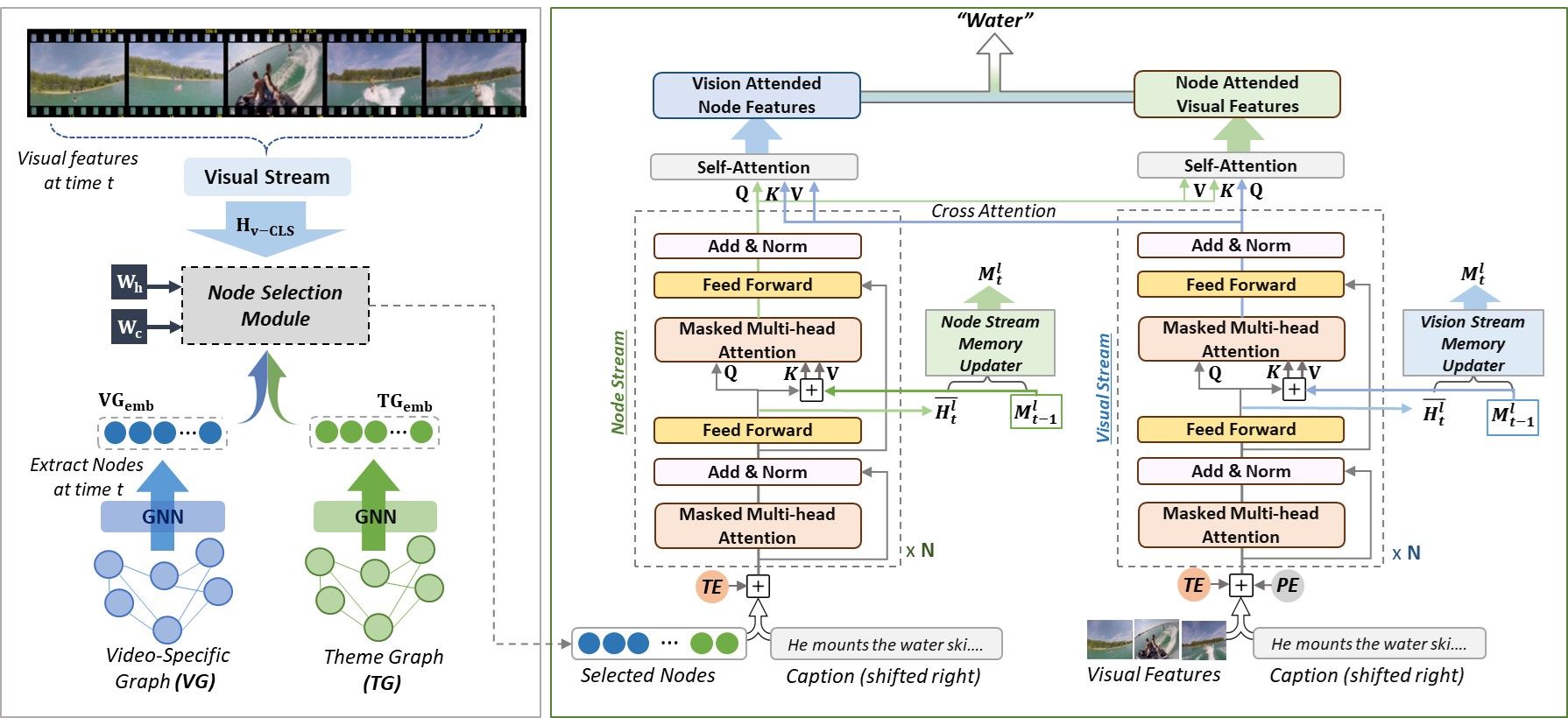}
  \caption{Architecture of GEM-VPC. At time $t$, the entire video-specific (VG) and theme graph (TG) corresponding to the action at time $t$ is fed into separate Graph Neural Networks. In the visual stream, visual features summed with positional (PE) and token type embeddings (TE) are inputted into a Recurrent Transformer and the sequence representation ($H_{\text{v-CLS}}$) is then used to select nodes from VG and TG in the node selection module. The selected nodes plus TE are fed into another Recurrent Transformer in the node stream. Cross-attention is employed between the visual and node stream and cross-attended features are finally fed into an MLP to predict the next word.}
  \label{fig: model_architecture}
\end{figure*}

\subsection{Theme Graph Creation}\label{tg_graph_creation}
We also create a theme graph for each action class to incorporate corpus-level information. Given an action predicted at $e_{vt}$, we collect the corresponding ground-truth training sentence at $e_{vt}$ and tag the nouns, verbs and adverbs to build a vocabulary for each action class. With the ASR-method, the action classes are created by the $k$-means algorithm to cluster the text embeddings of the action nodes. We retain the top-$n$ most frequent words for each action class vocabulary and following \cite{yao2019graph}, the individual words are connected based on word co-occurrence statistics to form a graph. 

\begin{equation}
    \text{PMI}(i,j) = \text{log}\frac{p(i,j)}{p(i)p(j)}
\end{equation}

\begin{equation}
    \text{NPMI} = \frac{\text{PMI}}{-\text{log}(p(i,j))}
\end{equation}

We utilise the normalised point-wise mutual information score (NPMI), where a positive score implies high semantic correlation between words. Here, $p(i,j) = \frac{\text{\#}S(i,j)}{\text{\#}S}$, $p(i) = \frac{\text{\#}S(i)}{\text{\#}S}$ and $p(j) = \frac{\text{\#}S(j)}{\text{\#}S}$ where $\text{\#}S(i)$ is the number of sentences in the corpus that contain word $i$, $\text{\#}S(i,j)$ is the number of sentences that contain both words and $\text{\#}S$ is the number of sentences in the corpus. For the corpus, we use the ground-truth sentences from external datasets (see Section \ref{Evaluation_Setup}). A word-to-word connection is made only if the NPMI score exceeds 0.10. A theme graph example is in Appendix F.

\subsection{VPC Model} \label{vpc_model}
GEM-VPC (Figure \ref{fig: model_architecture}) adopts a transformer-based shared encoder-decoder augmented with an external memory module to model temporal dependencies between events.
\\
\\
\noindent\textbf{1) Visual Stream}: 
At each timestep $t$ related to event $e_t$, we concatenate the visual features $F_V$ and predicted video captions $F_C$ from $e_t$. A [CLS] token is also prepended to learn the sequence representation. We denote the concatenated sequence as $F_{VC} = concat(F_V, F_C)$. $F_{VC}$ is fed into a transformer with learnt positional and token type embeddings (for indicating the token's modality type), which applies multi-head self attention (MHA): 

\begin{equation}
    \text{MHA}(Q,K,V) = \text{softmax}(\frac{QK}{\sqrt{d_k}} + M)V
\end{equation} \label{eq1}

\noindent where $Q = XW^Q$, $K = XW^K$, $V = XW^V$, $W^Q$, $W^K$, and $W^V$ are learnable parameters, $X = F_{VC}$ and $M$ is a masked matrix to prevent the model from attending to future words. The outputted intermediate hidden state $\bar{H}^l_t$ is then fed into another attention layer that performs MHA between $\bar{H}^l_t$ and past memory states for capturing history information.
\\
\\
\noindent\textbf{2) Node Stream}:
For each event (timestep), a representative action is extracted by using the predicted action label with the highest confidence score out of the predicted actions from $e_{vt}$. The matching theme graph for that action class is then obtained and fed through a Graph Attention Network (GAT) to learn theme node embeddings. For encoding the video-specific graph information, we feed the entire graph into another GAT and extract the node embeddings corresponding to timestep $t$. We denote the theme and video-specific graph node embeddings as $TG_{emb} \in \mathbb{R}^{N \times d}$ and $VG_{emb} \in \mathbb{R}^{M \times d}$, where $N$, $M$ are the number of nodes and $d$ is the embedding dimension. Specifically, we compute:

\begin{equation}
    H_\text{v-CLS} = \text{VisualStream}(F_{VC}) 
\end{equation}
\begin{equation}
    p_s = \text{softmax}(W_hH_\text{v-CLS}^t)^TW_cN_{emb}
\end{equation}

\noindent where $H_\text{v-CLS}$ is the [CLS] representation from the visual stream at time $t$, $W_h$ and $W_c$ are learnable, $N_{emb}$ is either $TG_{emb}$ or $VG_{emb}$ and $p_s$ contains probability scores for each node. The top-$n$ nodes yielding the highest probabilities from each $TG_{emb}$ and $VG_{emb}$ are then selected to be inputs for the node stream. Finally, we concatenate the selected nodes $F_N$ with the predicted captions $F_C$ and feed $F_{NC} = concat(F_N, F_C)$ through another transformer analogous to the one used in the visual stream. We do not add positional embeddings here as the selected nodes have no temporal order.
\\
\\
\noindent\textbf{3) Decoding the Caption}
Visual and node streams exchange information with cross attention: 

\begin{equation}
    H_\text{v-CA}, H_\text{n-CA} = \text{CrossAttention}(H_v, H_n)
\end{equation}

\noindent Here, $H_v$ and $H_n$ are the outputs from the visual and node stream respectively at time $t$ while $H_\text{v-CA}$ and $H_\text{n-CA}$ are node attended visual features and visual attended node features respectively. The concatenation of $H_\text{v-CA}$ and $H_\text{n-CA}$ is finally fed into a linear (MLP) layer and the next word predicted word is the $argmax$ of the output.
\\
\\
\noindent\textbf{4) Encoding Recurrence}
To capture temporal dependencies between events from previous timesteps, recent methods for encoding recurrence into transformer models are adopted for our visual and node stream. 
\textbf{A) MART:} memory augmented recurrent transformer \cite{lei2020mart}, using multi-head attention to encode the memory state. Given the intermediate hidden state $\bar{H}^l_t$, the memory updated intermediate hidden state $H^l_t$ is computed: 

\begin{equation}
    H^l_t  = \text{MHA}(M^l_{t-1}, \bar{H}^l_t, \bar{H}^l_t)
\end{equation}

\noindent where $M_{t-1}$ is the past memory calculated by:

\begin{equation}
    C^l_t = \text{tanh}(W^l_{mc}M^l_{t-1} + W^l_{sc}S^l_t + b^l_c)
\end{equation}
\begin{equation}
    Z^l_t = \text{sigmoid}(W^l_{mz}M^l_{t-1} + W^l_{sz}S^l_t + b^l_z)
\end{equation}
\begin{equation}
    M^l_t = (1-Z^t_l) \otimes C^l_t + Z^l_t \otimes M^l_{t-1}
\end{equation}

\noindent where $\otimes$ is the Hadamard product, $W^l_{mc}$, $W^l_{sc}$, $W^l_{mz}$, $W^l_{sz}$ are trainable weights, $b^l_c$ and $b^l_z$ are trainable bias, $C_t^l$ is the internal cell state and $Z^l_t$ is the update gate that controls which information to retain from previous memory states. 
\textbf{B) TinT:} proposed by \cite{yamazaki2023vltint}, utilising Hybrid Attention Mechanism (HAM) \cite{vo2021aei} to select information from previous hidden states:

\begin{equation}
    M^l_t = [M^l_{t-1};\bar{H}^l_t]
\end{equation}
\begin{equation}
    Z^l_t = HAM(M^l_{t-1}, \bar{H}^l_t)
\end{equation}
\begin{equation}
    H^l_t = \text{MLP}(mean(Att([\bar{H}^l_t; Z^l_t]))) + \bar{H}^l_t
\end{equation}

\noindent Here, `;' denotes concatenation along a new dimension, $mean(Att((\cdot))$ is self-attention applied on the new dimension and reduced by the mean operation, $M_{t}$ is the memory information at time $t$ ($M^l_0 = \emptyset$) and $\bar{H}^l_t$ is defined as above. 

\section{Evaluation Setup\protect\footnote{I\MakeLowercase{mplementation details can be found in \MakeUppercase{A}ppendix \MakeUppercase{G}}}}\label{Evaluation_Setup} 

\subsection{Datasets}
\noindent \textbf{1) ActivityNet Captions} \cite{krishna2017dense} consists of 10,009 training and 4,917 validation videos on people performing complex activities. On average, each video contains 3.65 event segments covering 36 seconds. We follow previous works \cite{lei2020mart} and split the original validation set into \textit{ae-val} and \textit{ae-test}. 
\\
\\
\noindent \textbf{2) YouCook2} \cite{zhou2018end} is for dense video procedural captioning in the recipe domain. It contains 1,333 training and 457 validation samples comprised specifically of instructional cooking videos. On average, videos are 5.26 minutes long with 7.7 event segments and each annotation for an event is a language description of the procedure's step covering 1.96 seconds. We report our results on the validation set (`\textit{yc2-val}').
\\
\\
\noindent \textbf{3) RecipeNLG} \cite{bien2020recipenlg} is for recipe generation, consisting of 2,231,142 cooking recipes and food entities from the recipes extracted using Named Entity Recognition. We use RecipeNLG as a supporting dataset to compute the NPMI scores when constructing the theme graphs for the YouCook2.

\begin{table*}[t]
  \centering
  
  \resizebox{1\linewidth}{!}{
  \begin{tabular}{@{}l||c|l|l|c|cccc|c}
    \toprule
     \multicolumn{9}{c}{\textit{\textbf{ae-test}}}  \\ \hline
     \textbf{Model} & \textbf{Conference} & \textbf{Year} & \textbf{Modalities} & \textbf{Integration Method} & \textbf{B4} $\uparrow$ & \textbf{M} $\uparrow$ & \textbf{C} $\uparrow$  & \textbf{R} $\uparrow$ & \textbf{R4} \\ \midrule
     VTrans \cite{zhou2018end} & CVPR & 2018 & V+F & Concatenation & 9.31 & 15.54 & 21.33 & 28.98 & - \\
     Trans-XL \cite{dai2019transformer} & ACL & 2019 & V+F & Concatenation & 10.25 & 14.91 & 21.71 & 30.25 & 8.54\\
     MDVC \cite{iashin2020multi} \dag & CVPR & 2020 & V+S+A & Concatenation & 8.50 & 14.28 & 17.57 & 25.48  & - \\
     BMT \cite{iashin2020better} \dag & BMVC & 2020 & V+A & CM Attention & 8.42 & 14.08 & 15.41 & 25.44  & - \\
     MART \cite{lei2020mart} & ACL & 2020 & V+F & Concatenation & 9.78 & 15.57 & 22.16  & - & 5.44\\
     MART-COOT \cite{ging2020coot} & NeurIPS & 2020 & V+L & Joint CM Space & 10.85 & 15.99 & 28.19  & - & -\\
     Trans-XLRG \cite{lei2020mart} & ACL & 2020 & V+F & Concatenation & 8.85 & 10.07 & 14.58 & 20.34  & - \\
     Motion-Aware \cite{hu2023motion} & ICASSP & 2023 & V+O & CM Attention & 11.90 & 16.54 & 30.13 & -  & 4.12 \\ 
     Memory Trans. \cite{song2021towards} & CVPR & 2021 & V+F & Concatenation & 11.74 & 15.64 & 26.55  & - & \textbf{2.75} \\
     VLCAP \cite{yamazaki2022vlcap} & ICIP & 2022 & V+L & CM Attention & 13.38 & 17.48 & 31.29  & 35.99 & 4.18 \\
     VLTinT w/ CL \cite{yamazaki2023vltint} & AAAI & 2023 & V+L+O & CM Attention & \underline{14.50} & \underline{17.97} & 31.13 & \textbf{36.56} & 4.75 \\ 
     VLTinT w/ CL$^\ast$  \cite{yamazaki2023vltint} & AAAI & 2023 & V+L+O  & CM Attention & 14.32 & 17.84 & \underline{31.83} & \underline{36.51} & 5.16 \\ 
     VLTinT w/o CL \cite{yamazaki2023vltint} & AAAI & 2023 & V+L+O & CM Attention & 13.80 & 17.72 & 30.59 & 36.11 & - \\
     VGCSN+CHPG \cite{yu2024exploring} & ICASSP & 2024 & V+L+O+C & CM Attention & 11.80 & 16.51 & 29.69 & - & \underline{4.02} \\ 
     \hline 
     GEM-VPC w/ No Recurrence & - & 2024 & V+G(V+A+C) & CM Attention & 12.82 & 17.4 & 26.97 & 33.45 & 7.28 \\
     GEM-VPC w/ MART decoder & - & 2024 & V+G(V+A+C) & CM Attention & 13.47 & 17.38 & 30.38 & 35.8 & 5.93 \\
     GEM-VPC w/ TinT decoder & - & 2024 & V+G(V+A+C) & CM Attention & \textbf{14.54} & \textbf{17.99} & \textbf{32.62} & \underline{36.51} & 5.17 \\
    \bottomrule
  \end{tabular}}
  \caption{Automatic scores for ActivityNet \textit{ae-test}. In `Modalities', V=visual, F=optical flow, O=bounding box object visual features, A=audio, S=speech, L=language, G(V+A+C)=graph built with visual, audio modality and commonsense. \dag \space indicates results computed by ourselves. $\ast$ are computed by rerunning the model with our own environment. `Integration Method'=how to integrate the distinct modalities (see Appendix H for specific meanings).}\label{tab: automatic_metrics_anet_test}
\end{table*}

\begin{table*}[t]
  \centering
  
  \resizebox{1\linewidth}{!}{
  \begin{tabular}{@{}l||c|l|l|c|c|cccc|c}
    \toprule
     \multicolumn{9}{c}{\textit{\textbf{yc2-val}}}  \\ \hline
     \textbf{Model} & \textbf{Conference} & \textbf{Year} & \textbf{Modalities} & \textbf{Pretraining} & \textbf{Integration Method} & \textbf{B4} $\uparrow$ & \textbf{M} $\uparrow$ & \textbf{C} $\uparrow$  & \textbf{R} $\uparrow$ & \textbf{R4} \\ \midrule
     VTrans \cite{zhou2018end} & CVPR & 2018 & V+F & \ding{55} & Concatenation & 7.62 & 15.65 & 32.26 & - &  7.83  \\
     Trans-XL \cite{dai2019transformer} & ACL & 2019 & V+F & \ding{55} & Concatenation & 6.56 & 14.76 & 26.35 & - &  6.30 \\
     MART \cite{lei2020mart} & ACL & 2020 & V+F & \ding{55} & Concatenation & 8.00 & 15.90 & 35.74& - &  \underline{4.39} \\
     MART-COOT \cite{ging2020coot} & NeurIPS & 2020 & V+L & \ding{55} & Joint CM Space & 9.44 & 18.17 & 46.06 & - &  6.30 \\
     Trans-XLRG \cite{lei2020mart} & ACL & 2019 & V+F  & \ding{55} & Concatenation & 6.63 & 14.74 & 25.93 & - &  6.03 \\
     VLCAP \cite{yamazaki2022vlcap} & ICIP & 2022 & V+L & \ding{55} & CM Attention & 9.56 & 17.95 & 49.41  & 35.17 & 5.16 \\
     VLTinT \cite{yamazaki2023vltint} & AAAI & 2023 & V+L & \ding{55}  & CM Attention & 9.40 & 17.94 & 48.70 & 34.55 & \textbf{4.29} \\ 
     VGCSN+CHPG \cite{yu2024exploring} & ICASSP & 2024 & V+L+O+C & \ding{55} & CM Attention & 6.80 & 14.50 & 27.21  & - & - \\
     DECEMBERT \cite{tang2021decembert} & NAACL & 2021 & V+L+S  & \ding{51} & CM Pretraining & \textbf{11.92} & \textbf{20.01} & \underline{58.02} & \textbf{40.22} & - \\
     MTrans+COOT+MIL-NCE PT \cite{tang2021decembert} & NAACL & 2021 & V+L & \ding{51}  &  Joint CM Space & 11.05 & 19.79 & 55.57 & 37.51 & - \\
     MART+COOT+MIL-NCE PT\cite{tang2021decembert} & NAACL & 2021 & V+L  & \ding{51} & Joint CM Space & 11.30 & 19.85 & 57.24 & \underline{37.94} & - \\ \hline 
     GEM-VPC w/ No Recurrence & - & 2024 & V+G(S+A+C) & \ding{55} & CM Attention & 11.03 & \textbf{20.01} & \textbf{58.49} &  36.89 & 4.64  \\
     GEM-VPC w/ MART decoder & - & 2024 & V+G(S+A+C) & \ding{55} & CM Attention & 11.01 & \underline{19.86} & 54.84 &  36.81 & 4.47 \\
     GEM-VPC w/ TinT decoder & - & 2024 & V+G(S+A+C) & \ding{55} &  CM Attention & \underline{11.47} & 19.72 & 56.00 & 37.48 & 4.91 \\ \bottomrule 
    \end{tabular}}
  \caption{Automatic scores for baselines and GEM-VPC on YouCook2. The `Modalities' and `Integration Method' columns are the same as Table \ref{tab: automatic_metrics_anet_test}. Additionally, `G(S+A+C)' is graph built with speech/audio modality and commonsense, `CM Pretaining' indicates the use of pretraining objectives like masked language modelling. The `Pretraining' column indicates whether the model has been pretrained on an external video dataset. }\label{tab: automatic_metrics_yc2}
\end{table*}

\subsection{Evaluation Metrics}
We follow previous VPC works and evaluate with: BLEU-4 (B4) \cite{papineni2002bleu}, METEOR (M) \cite{banerjee2005meteor}, CIDEr (C) \cite{vedantam2015cider}, and ROUGE-L (R) \cite{lin2004rouge}. We also analyse the repetitiveness and diversity of the captions by measuring 2-gram diversity (Div2) \cite{shetty2017speaking} and 4-gram repetition \cite{xiong2018move}.

\section{{Results}\protect\footnote{Appendix J shows qualitative examples of generated captions from our model versus state-of-the-art}}
\subsection{Performance Against SOTA} \label{performance_sota}
We compare GEM-VPC with prior SOTA on ActivityNet Captions's \textit{ae-test} split (Table \ref{tab: automatic_metrics_anet_test}) and YouCook2's validation split (Table \ref{tab: automatic_metrics_yc2}). 

Our best model (GEM-VPC w/ TinT decoder) evidently outperforms most of the existing baselines. VLTinT w/ CL and w/o CL is the VLTinT model trained with their novel contrastive loss (in addition to the classic MLE loss) and without their contrastive loss respectively. Specifically, GEM-VPC w/ TinT decoder outperforms VLTinT w/ CL on BLEU-4, METEOR and CIDEr and all metrics when considering the VLTinT w/o CL variant which is optimised using the same MLE loss as our model. For a more accurate comparison, we rerun VLTinT w/ CL (with their optimal parameters) in our own environment and record the results under VLTinT w/ CL$^\ast$. As shown, GEM-VPC w/ TinT decoder yields higher BLEU-4, METEOR and CIDEr scores than VLTinT w/ CL$^\ast$ with similar ROUGE and R4. While R4 does not outperform some baselines, the lower repetition does not necessarily mean good caption quality as lower repetition can be simply achieved by generating words unrelated to the video content. Hence, a strong model should have a balance of high $n$-gram metrics and a low repetition score. Examining YouCook2, our model variants achieve higher $n$-gram scores with relatively low repetition of 4.6-4.9 compared to baselines with no pretraining (first 6 baselines). Even when comparing with the last 3 baselines with pretraining methods and a large separate instructional video dataset (HowTo100M \cite{miech2019howto100m}), we achieve similar scores with our best CIDEr score (58.49) outperforming all baselines. 

\subsection{Ablation Studies}
\noindent\textbf{Different Input Modalities}: Our model is examined with different modality settings in Table \ref{tab: anet_yc2_ablation}. Using visual features alone (Exp \# \textcircled{1}) for both datasets yields the worst performance with the lowest scores across all $n$-gram metrics. Using nodes only (Exp \# \textcircled{2}) can substantially improve the scores, although this produces higher repetition and lower diversity. We also find that the setting using visual features combined with node features results in significant performance improvement across all metrics (Exp \# \textcircled{5}). Comparing \textcircled{3} and \textcircled{4}, inputting visual+VG features exhibit higher $n$-gram metrics than using visual+TG features for both datasets, indicating that the VG provide more useful information representative of the video content. R4 and Div2 scores remain similar for ActivityNet, but that for YouCook2 yields lower repetition/higher diversity. However as previously noted, lower repetition/higher diversity does not mean good caption quality if the $n$-gram metrics are also low. Overall, we show that incorporating video-specific information and the TG corpus-level information (Exp \# \textcircled{5}) is superior. We further experimented by adding a separate stream to process the raw audio features. Comparing \textcircled{5} and \textcircled{6} for both datasets, adding audio information slightly improves the repetition/diversity at the cost of lower B4 and CIDEr. This could be due to a misalignment in the audio track and the video's topic e.g. there are cases where users upload background music unrelated to the video contents. Moreover, we examine noisy background audio that could potentially confuse the model. For YouCook2, by examining unprocessed speech features (Exp \# \textcircled{7}), inputting the visual and speech transcript can produce competitive performance. However, this can be further enhanced by incorporating node information from VG and TG as seen in \textcircled{8} which yields the highest B4 and CIDEr out of all the settings whilst maintaining competitive Div2 and R4. 
\begin{table}[t]
  \centering
  \resizebox{1\linewidth}{!}{
  \begin{tabular}{@{}c||ccccc||ccc|cc}
    \toprule
     \multicolumn{11}{c}{\textbf{ActivityNet (\textit{ae-test})}} \\ \hline
     \textbf{Exp $\#$} & \textbf{V} & \textbf{VG} & \textbf{TG} & \textbf{A} & \textbf{S} & \textbf{B4} $\uparrow$ & \textbf{M} $\uparrow$ & \textbf{C} $\uparrow$  & \textbf{Div2} $\uparrow$ &\textbf{R4} $\downarrow$ \\ \midrule
     \textcircled{1} & \ding{51} & \ding{55} & \ding{55} & \ding{55} & \ding{55} & 12.90 & 16.92 &  28.27 & 75.65 & 6.00 \\
     \textcircled{2} & \ding{55} & \ding{51} & \ding{51} & \ding{55} & \ding{55} & 10.63 & 16.51 &  20.75 & 74.83 & 7.66 \\
    \textcircled{3} & \ding{51} & \ding{51} & \ding{55} & \ding{55} & \ding{55} & \underline{13.27} & 17.24 &  28.99 & 74.29 & 6.93 \\
     \textcircled{4} & \ding{51} & \ding{55} & \ding{51} & \ding{55} & \ding{55}  & 13.12 & 17.09 &  27.97 & 75.02 & 7.01 \\
    \textcircled{5} & \ding{51} & \ding{51} & \ding{51} & \ding{55} & \ding{55} & \textbf{13.47} & \underline{17.38} &  \textbf{30.38} & \underline{75.74} & \underline{5.93} \\
    \textcircled{6} & \ding{51} & \ding{51} & \ding{51} & \ding{51} & \ding{55} & 13.16 & \textbf{17.40} &  \underline{29.88} & \textbf{76.24} & \textbf{5.80} \\ \hline
    \multicolumn{11}{c}{\textbf{YouCook2} \textbf{(\textit{yc2-val)}}} \\ \hline
    \textcircled{1} & \ding{51} & \ding{55} & \ding{55} & \ding{55} & \ding{55} & 7.12 &  15.25 & 30.12 & \underline{70.75} & 3.66 \\
     \textcircled{2} & \ding{55} & \ding{51} & \ding{51} & \ding{55} & \ding{55} & 9.91 &  18.65 & 44.50 & 65.38 & 6.33\\
    \textcircled{3} & \ding{51} & \ding{51} & \ding{55} & \ding{55} & \ding{55} & 10.82 & 19.42 & 54.73 & 67.11 & 4.65 \\
     \textcircled{4} & \ding{51} & \ding{55} & \ding{51} & \ding{55} & \ding{55} & 8.06 &  16.35 & 36.68 & 69.96 & 3.72 \\
    \textcircled{5} & \ding{51} & \ding{51} & \ding{51} & \ding{55} & \ding{55} & \underline{11.03} &  \textbf{20.01} & \underline{58.49} & 67.08 & 4.64\\
    \textcircled{6} & \ding{51} & \ding{51} & \ding{51} & \ding{51} & \ding{55} & 9.73 &  18.50 & 53.33 & 68.22 & 4.36 \\
    \textcircled{7} & \ding{51} & \ding{55} & \ding{55} & \ding{55} & \ding{51} & 10.94 &  19.90 & 57.45 & \textbf{71.55} & \textbf{1.94} \\
    \textcircled{8} & \ding{51} & \ding{51} & \ding{51} & \ding{55} & \ding{51} & \textbf{11.56} & \underline{19.98} & \textbf{58.70} & 70.46 & \underline{2.61} \\
    \bottomrule
  \end{tabular}}
  \caption{GEM-VPC performance with different input modalities. ActivityNet (with MART decoder); YouCook2 (with No Recurrence setting). Exp \# is the experiment number and V, VG, TG, A, S stand for visual, video-specific graph, theme graph, audio and speech.}\label{tab: anet_yc2_ablation}
\end{table}

\begin{figure}%
    \centering
    \subfloat\centering{{\includegraphics[width=0.22\textwidth]{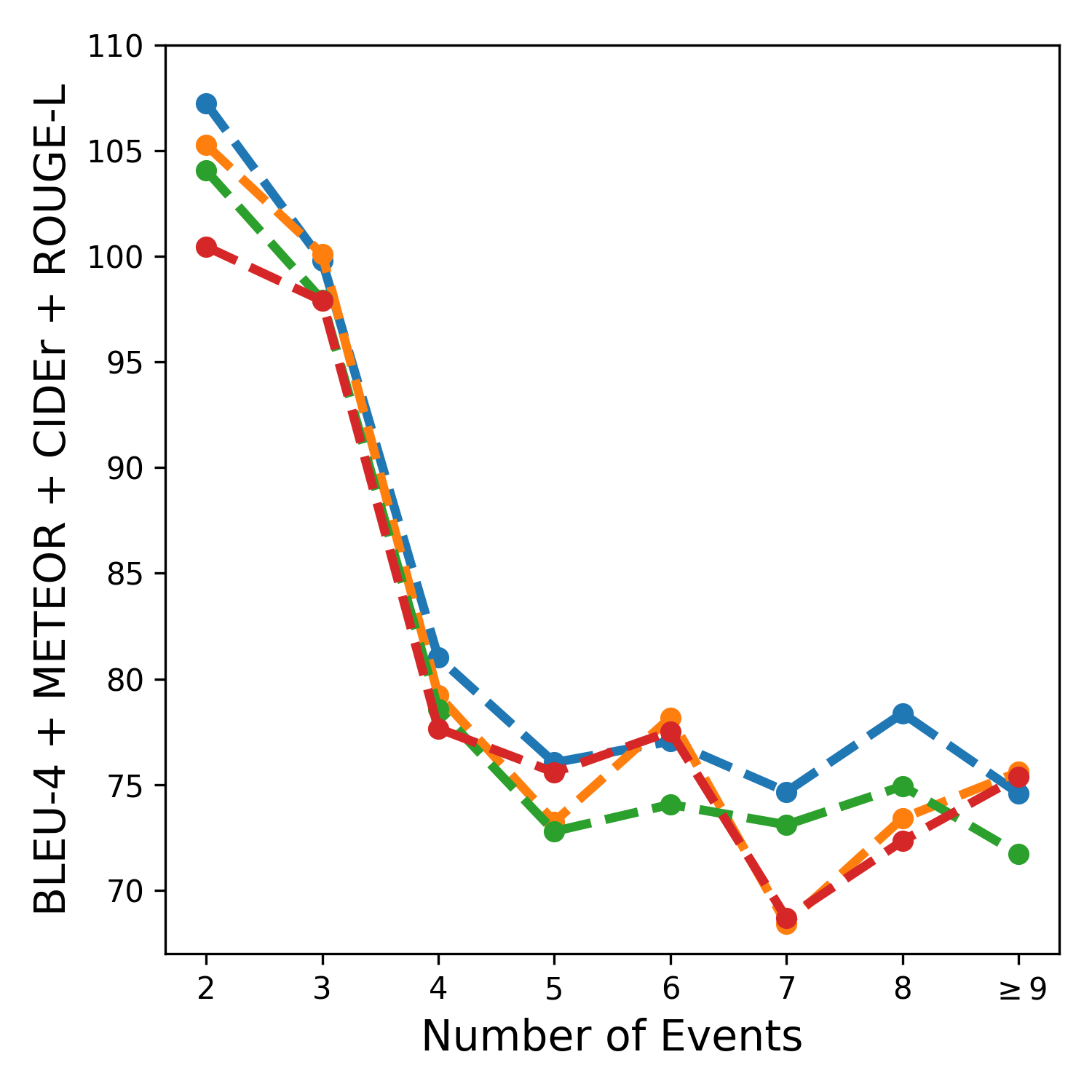} }}%
    \hspace{0em}
    \subfloat\centering{{\includegraphics[width=0.22\textwidth]{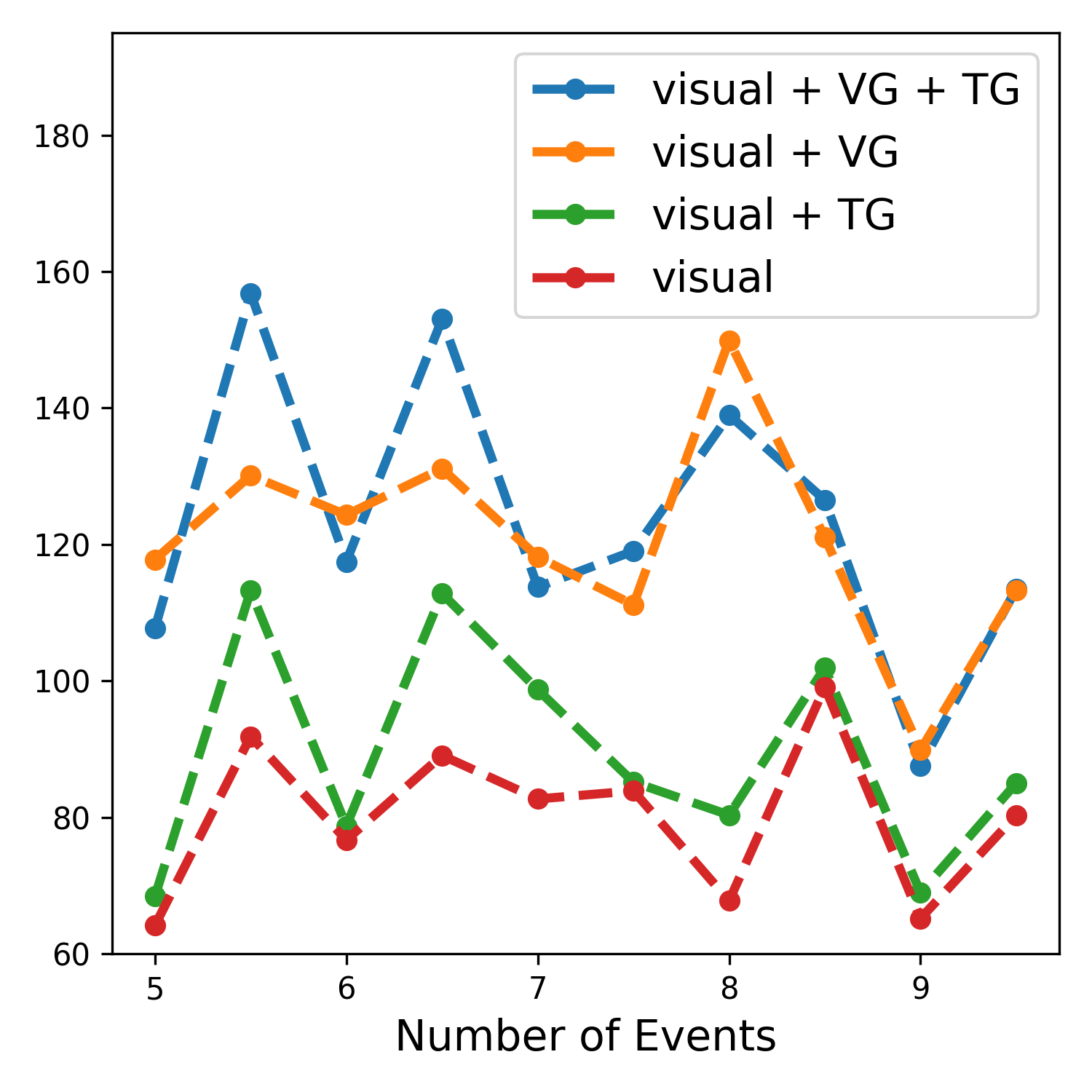} }}%
   \caption{Sum of $n$-gram metrics on ActivityNet (\textit{ae-val}+\textit{ae-test}) (left) and YouCook2 (\textit{yc2-val}) (right) across samples with different number of events.}\label{fig:num_events_analysis}
\end{figure}



\textbf{Different Decoders}: We evaluated different methods for encoding recurrence using MART, TinT, and a `No Recurrence' setting, as in the last three rows of Table \ref{tab: automatic_metrics_anet_test}/\ref{tab: automatic_metrics_yc2} for ActivityNet/YouCook2. For ActivityNet, the TinT decoder achieved the best results across all metrics, followed by MART, with the No Recurrence setting performing the worst, indicating the importance of a recurrent memory module. Conversely, YouCook2 results showed that the No Recurrence setting yielded the highest METEOR (20.0) and CIDEr (58.5) scores, while TinT improved BLEU-4 and ROUGE-L but had the lowest METEOR and R4. This suggests that encoding recurrence benefits captioning if the current timestep relies on past information. We analysed samples by their total timesteps and plotted the average sum of $n$-gram metrics for each group in Figure \ref{fig:num_events_analysis}. For YouCook2, even without recurrence, decoding captions for samples with more timesteps wasn't necessarily more complex. However, for ActivityNet, scores decreased with more timesteps, highlighting the need for recurrent information. This aligns with the MART paper's findings on ActivityNet, though YouCook2 wasn't tested in their study \cite{lei2020mart}.

\subsection{Qualitative Examples} \label{christmas example}
\begin{figure}[H]
  \centering
  \includegraphics[width=1\linewidth]{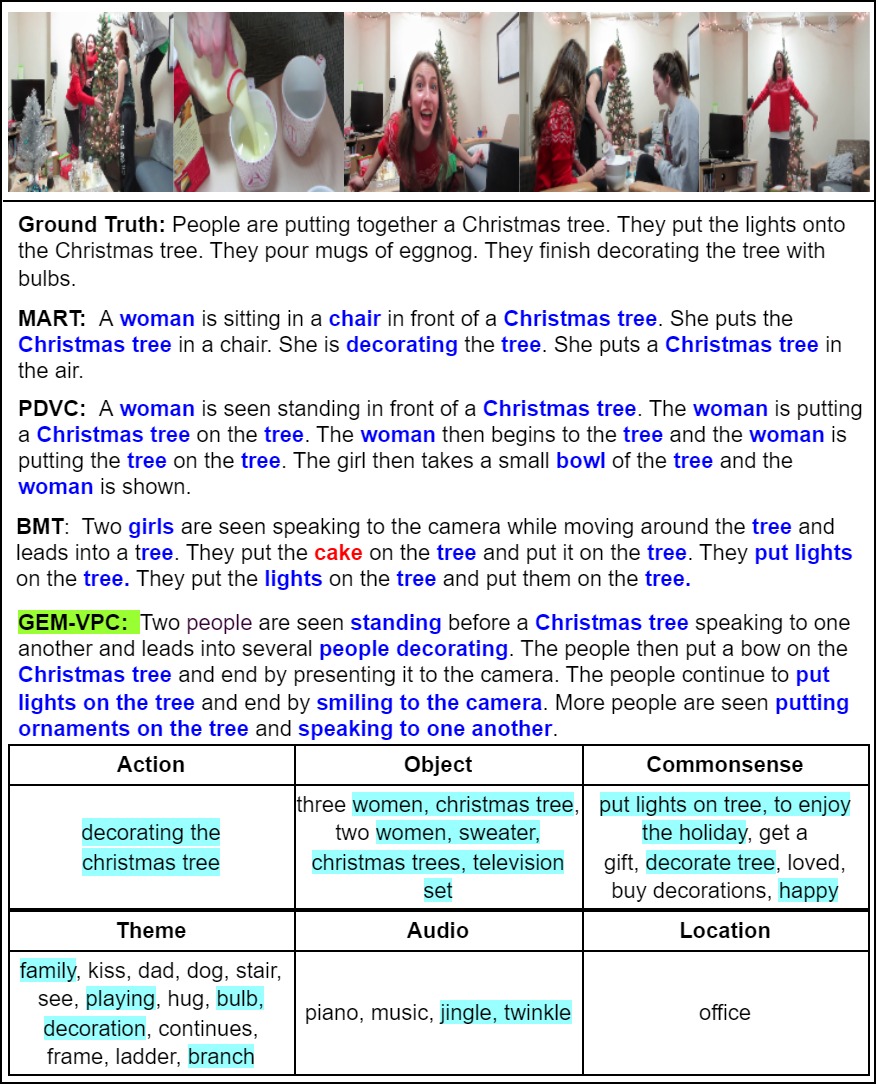}
  \caption{Qualitative Example from ActivityNet. Blue words in the machine-generated captions are visually grounding to the video, while red words represent irrelevant words that are `hallucinated' by the model.}
  \label{fig: christmas_example}
\end{figure}

Qualitative Examples for the start-of-the art methods versus GEM-VPC are shown in Figure \ref{fig: christmas_example} and Appendix J. We collect the top-10 selected nodes by confidence score at each timestep during inference and display the selected nodes and their types in the table after each example. Highlighted blue words indicate information related to the video's theme. Evidently, the commonsense-enhanced video graph and theme graph assists our model in producing concepts and phrases relevant to the video segment. For instance in Figure \ref{fig: christmas_example}, GEM-VPC mentions relevant phrases like `\textit{smiling to the camera}' and `\textit{putting ornaments on the tree}' which were perhaps derived from selected nodes such as `\textit{happy}', `\textit{decoration}' and `\textit{jingle}'. Conversely, other baselines (see Appendix J) will sometimes mention irrelevant concepts such as in the last instance where BMT incorrectly outputs `\textit{brushing his face}' in contrast to our model which is capable of recognising the action of a person shaving his beard. 

\section{Conclusion}
We introduced GEM-VPC, a novel framework for video captioning (VPC) that leverages multimodal information and external knowledge. We construct a commonsense-enhanced video-specific graph for key events and context, and a theme graph from ground-truth captions to represent word relationships. These graphs are processed by separate GNNs, and a node selection module identifies useful nodes for caption decoding. The selected nodes and supporting information (visual, audio, etc.) are fed into a transformer with multiple streams for different modalities, followed by a cross-attention module for inter-stream information exchange. Experiments on benchmark datasets demonstrate that GEM-VPC outperforms existing baselines, generating coherent and visually-grounded captions.

{\small
\bibliographystyle{ieee_fullname}
\bibliography{egbib}
}
\newpage
\appendix
\section{Video-Specific Graph Statistics} \label{graph_stats}
Total count for the different node types for the ActivityNet and YouCook2 video-specific graphs. On average, the ActivityNet and YouCook2 graphs have 57.04 and 127.83 nodes respectively with the largest graphs containing 259 and 304 nodes respectively. 
\raggedbottom
\begin{table}[H]
  \centering
  \resizebox{0.9\linewidth}{!}{
  \begin{tabular}{@{}l||cc}
    \toprule
     \textbf{Node Type} & \textbf{ActivityNet} & \textbf{YouCook2}  \\ \midrule
    Action  & 74,017 & 43,555 \\
    Location & 54,802 & - \\
    Contextual Phrase & - & 13,464 \\
    Object & 198,905 & 64,379  \\
    Audio & 49,496 & 2,848 \\
    Commonsense & 472,534 & 97,147 \\  
    \bottomrule
  \end{tabular}}
  \caption{Count of different node types for both the ActivityNet and YouCook2 video-specific graphs.} \label{tab: node_type_count} 
\end{table}

\section{Node Type Importance}
To examine the importance of different node types in the video-specific graph, for each video sample during inference, we extract the top-10 selected nodes chosen by our node selection module at each event timestep. Then, for each node type, we count the frequency of selected nodes across the timesteps and divide the count by the total number of nodes (of that same type) that are present in the video-specific graph to get the proportion (normalised count). Finally, the proportions computed from each video sample is averaged across the validation set. This number reflects the expected probability that a node of a specific type will be selected to be part of the top-10 nodes. The results for both datasets is shown in the table below. For example, 48.99\% in the table means that for ActivityNet video-specific graphs, if the node is an action node, then it has 48.99\% chance to be in the top-10 nodes as ranked by our node selection module. 

\begin{table}[H]
  \centering
  \resizebox{1\linewidth}{!}{
  \begin{tabular}{@{}l||cc}
    \toprule
     \textbf{Node Type} & \textbf{ActivityNet (\%)} & \textbf{YouCook2 (\%)}  \\ \midrule
    Action  & 48.99 & 51.48 \\
    Location & 54.30 & - \\
    Contextual Phrase & - & 55.56\\
    Object & 39.79 & 43.61 \\
    Audio & 54.24 & 59.76 \\
    Commonsense &  30.80 & 10.74 \\  
    \bottomrule
  \end{tabular}}
  \caption{Average proportion of selected nodes for each node type in the video-specific graphs for both ActivityNet and YouCook2.} \label{tab: node_type_importance_analysis} 
\end{table}

\noindent Examining Table \ref{tab: node_type_importance_analysis}, for both ActivityNet and YouCook2, the node types with the highest average selected proportions were the location/contextual phrase, audio and action nodes, indicating that these node types tend to be more vital for video understanding. Action nodes having a high chance of being selected is not surprising, as this node captures information closely aligned with the VPC task where the aim is to generate captions describing the action and events in the video segment. Similarly for ActivityNet, the location nodes may be important as the action/events happening in the video are often closely related to location e.g. videos about water skiing often happen in locations with water. Moreover, for YouCook2, the contextual phrase nodes are most likely significant as they provide similar information to the action nodes. The large percentage of audio nodes selected for both datasets may be unexpected at first as raw video sounds tend to contain noisy background information. However, as mentioned in Section 3.1 of the main paper, we already perform extra post-processing in an attempt to retain only the relevant audio labels. For both datasets, the node type with the second least selection probability are object nodes with on average, 39-44\% considered as important. This is however still a relatively large proportion, suggesting that object nodes are significant for VPC. Finally, we observe that a majority of the commonsense nodes were not useful, especially for YouCook2, despite the large count of commonsense nodes in both graphs (see Appendix \ref{graph_stats}). This is perhaps attributed to the fact that Comet-ATOMIC2020 focuses on generating social commonsense such as people's reactions, intents and desires relating to a specific event. However, we find that the ground-truth captions are often limited in detail whereby annotators do not always describe such information but mainly just simply focus on stating what is visually happening in the video. Nevertheless, a relatively large proportion of 30.8\% is still selected from the ActivityNet video-specific graphs, suggesting that this social commonsense knowledge can still provide useful contextual cues for videos that are similar in nature to the ones in ActivityNet. 

\raggedbottom

\section{Number of Nodes Selected}
We report the performance of our model when changing the different maximum number of nodes that can be selected from each video-specific graph and each theme graph per timestep. Results for the ActivityNet and YouCook2 dataset are displayed in Table \ref{tab: anet_num_nodes_analysis} and Table \ref{tab: yc2_num_nodes_analysis} respectively. For ActivityNet, the best $n$-gram and repetition scores can be achieved when using 20 nodes (10 nodes selected from each of the video-specific and theme graphs at each timestep) or 40 nodes (20 nodes selected from each graph at each timestep). For YouCook2, we find that the best performance stabilises at around 60-80 total nodes. 

\begin{table}[H]
  \centering
  \resizebox{1\linewidth}{!}{
  \begin{tabular}{@{}c||cccc|cc}
    \toprule
     & \multicolumn{6}{|c}{\textbf{ActivityNet}} \\ \hline
     \textbf{\# Nodes} & \textbf{B4} $\uparrow$ & \textbf{M} $\uparrow$ & \textbf{C} $\uparrow$ & \textbf{R} $\uparrow$& \textbf{Div2} $\uparrow$& \textbf{R4} $\downarrow$\\ \midrule
    10  & 13.80 & 17.40 & 31.21 & 35.88 & 75.22 & 6.50\\
    20 & \textbf{13.91} & 17.40 & \textbf{31.45} & \textbf{36} & 75.19 & 6.41 \\
    40 & \textbf{13.91} & \textbf{17.47} & 30.68 & 35.97 & \textbf{75.75} &  \textbf{6.18} \\
    60 & 13.51 & 17.38 & 30.74 & 35.82 & 75.21 & 6.42\\
    \bottomrule
  \end{tabular}}
  \caption{Performance of our model by setting different maximum number of nodes that can be selected from our node selection module at each timestep on ActivityNet. All results are reported using the model w/ MART decoder with video and node input features.} \label{tab: anet_num_nodes_analysis} 
\end{table}

\begin{table}[H]
  \centering
  \resizebox{1\linewidth}{!}{
  \begin{tabular}{@{}c||cccc|cc}
    \toprule
     & \multicolumn{6}{|c}{\textbf{YouCook2}} \\ \hline
     \textbf{\# Nodes} & \textbf{B4} $\uparrow$ & \textbf{M} $\uparrow$ & \textbf{C} $\uparrow$& \textbf{R} $\uparrow$ & \textbf{Div2} $\uparrow$& \textbf{R4} $\downarrow$\\ \midrule
    10  & 10.39 & 18.82 & 52.24 & 35.52 & 65.87 & 5.47\\
    20 & 10.73 & 19.27 & 54.21 & 35.97 & 67.63 & 4.80 \\
    40 & 10.88 & 19.58 & 57.38 & 36.69 & 66.03 & 5.40 \\
    60 & 11.03 & 20.01 & \textbf{58.49} & \textbf{36.89} & 67.08 & \textbf{4.64}\\
    80 & \textbf{11.23} & \textbf{20.04} & 57.84 & 36.78 & \textbf{67.77} & 4.75\\
    \bottomrule
  \end{tabular}}
  \caption{Performance of our model by setting different maximum number of nodes that can be selected from our node selection module at each timestep on YouCook2. All results are reported using the model w/o Recurrence with video and node input features.} \label{tab: yc2_num_nodes_analysis} 
\end{table}

\begin{table*}[t]
  \centering
  \resizebox{1\linewidth}{!}{
  \begin{tabular}{@{}l|l|l}
    \toprule
    \textbf{Relation} & \textbf{Description} & \textbf{Example ($<$head$>$$<$\texttt{relation}$>$$<$tail$>$)}\\ \midrule
    ObjectUse  & describes everyday affordances or uses of objects & put into pan \texttt{ObjectUse} frying \\
    MadeUpOf  & describes a part, portion or makeup of an entity & making cake \texttt{MadeUpOf} eggs \\
    HasProperty  & describes entities’ general characteristics & boiling water \texttt{HasProperty} heat \\
    CapableOf  & describe abilities and capabilities of everyday living entities & cut cake \texttt{CapableOf} celebrate birthday \\
    isAfter & events that can follow an event & mop the floor \texttt{isAfter} sweep the floor \\
    HasSubEvent & provides the internal structure of an event & boil the dumplings \texttt{HasSubEvent} boils water \\
    isBefore & events that can precede an event & opens a gift \texttt{isBefore} rips wrapping paper \\
    xNeed & describes a precondition for an agent to achieve the event & give a gift \texttt{xNeed} buys the presents\\
    xAttr & describes personas or attributes perceived by others given an event &  decorates Christmas tree \texttt{xAttr} festive \\
    xEffect/oEffect & actions that happen to an agent that may occur after the event & gives a present \texttt{xEffect} gets thanked \\
    xReact/oReact & emotional reactions of participants in an event & gives a present \texttt{xReact} feels happy \\
    xWant/oWant & postcondition desires after an event & gives a present \texttt{xWant} wants to hug \\
    xIntent &  defines the likely intent of an agent & pour sauce  on food \texttt{xIntent} add flavour \\
    \bottomrule
  \end{tabular}}
  \caption{Relations in Comet-ATOMIC2020 used to generate the commonsense nodes for the video-specific graph and their corresponding descriptions. The `head' indicates the input phrase that is fed into Comet-ATOMIC2020 and the `tail' is the possible generated commonsense.} \label{tab: relation_description} 
\end{table*}

\section{Performance Across Different Video Categories}
To examine how our model performs across different types of videos, we compute the average sum of BLEU-4, METEOR, CIDEr and ROUGE-L across 14 different categories for the ActivityNet validation and testing split. These categories are provided by the user when uploading the video and roughly represent the video's main topic.  For this experiment, 3 different types of input modalities are tested: 1) using video visual features only (visual), 2) using visual features combined with node features chosen by the node selection module (visual + nodes), and 3) using visual features combined with node features and audio features (visual + nodes + audio). 

\begin{figure}[H]
  \centering
  \includegraphics[width=1\linewidth]{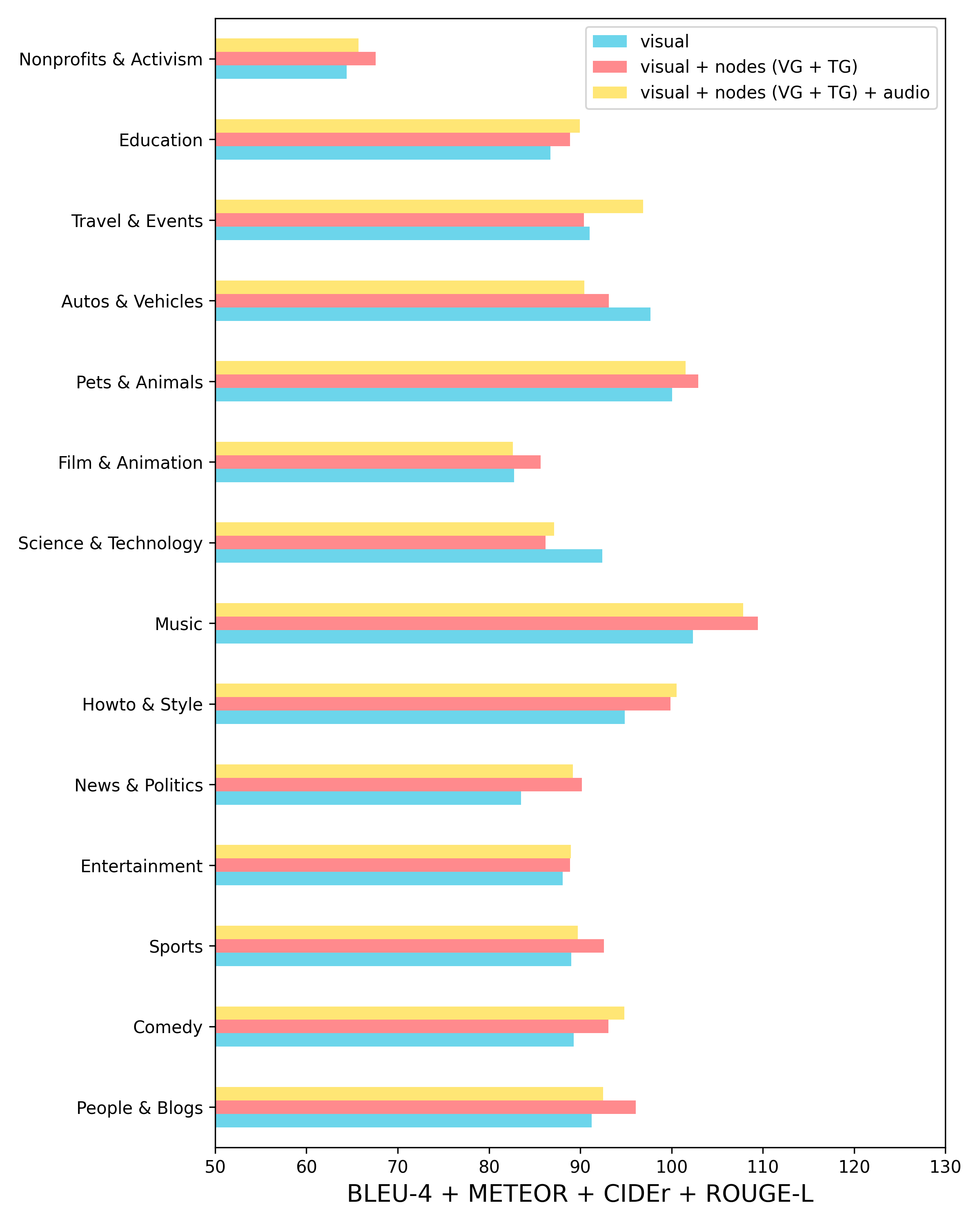}
  \caption{Sum of BLEU-4, METEOR, CIDEr and ROUGE-L scores for the ActivityNet predicted captions across the different video categories using 3 different input modalities (visual only, visual + nodes, visual + nodes + audio). The scores are obtained from the combined validation (\textit{ae-val}) and testing set (\textit{ae-test}).}
  \label{fig: anet_category_performance}
\end{figure}

Examining Figure \ref{fig: anet_category_performance}, when comparing video versus visual + nodes, we find that visual + nodes does better than visual only in all categories except for `Travel \& Events', `Autos \& Vehicles', and `Science \& Technology'. In particular, the largest gap occurs in the 2 latter categories. A reason for this may be due to a lack of action classes related to these categories in which the TimeSformer model is capable of predicting, which subsequently affects the quality of the nodes in the video-specific graph. For instance, there are no specific action classes that are related to `Science \& Technology' in the Kinetics600 dataset in which the TimeSformer model was trained on, while there are only 4 action classes that are related to `auto maintenance' (\textit{`changing oil'}, \textit{`changing wheel'}, \textit{`checking tires'}, \textit{`pumping gas'}). Furthermore, we observe that adding audio features to the model does not necessarily provide useful context cues for all categories. This is perhaps due to a misalignment between the audio track and video's topic. For example, people will often put a sound track with music even when the video itself is not about `Music'. However, we do find that audio helps in improving performance for categories related to `Education', `Travel \& Events', `Howto \& Style', and `Comedy'. 

In summary, visual + nodes performs the best in general, outperforming the other 2 model variants for 7 out of the 14 categories. This aligns with the findings from Section 5.2. Visual + nodes + audio is the second-best with superior performance in 5 categories. This is finally followed by the visual only setting, whereby visual features alone clearly does not provide enough contextual information to generate high quality captions and thus, only benefits 2 out of the 14 categories.

\section{Relation Description} \label{appendix: relation_description}
The relation tokens used to extract knowledge from the Comet-ATOMIC2020 neural knowledge model for the commonsense nodes in the video-specific graphs and their corresponding descriptions are detailed in Table \ref{tab: relation_description}. The descriptions are taken from the official Comet-ATOMIC2020 paper \cite{hwang2021comet}. For the ActivityNet graphs, all relations below were used except for \texttt{isAfter}, \texttt{isBefore}, \texttt{MadeUpOf}, \texttt{ObjectUse} and \texttt{HasProperty}. Although \texttt{isAfter} and \texttt{isBefore} relations may be useful, we find that the commonsense generated using these relations for the ActivityNet data tends to produce similar results to \texttt{xNeed} and \texttt{xEffect/oEffect} and so we disregard these relations to help reduce the number of commonsense nodes in the graphs. \texttt{MadeUpOf}, \texttt{ObjectUse} and \texttt{HasProperty} are further ignored as information about properties, compositions or characteristics of entities are not closely aligned with the content in the ActivityNet captions. For the YouCook2 graphs, all relations below were used except for \texttt{xReact/oReact}, \texttt{xAttr} and \texttt{xWant/oWant} as we believe information about an event's attributes and individual's subjective reactions/desires may not be useful for captioning instructional cooking videos. 

\section{Theme Graph Example} \label{appendix: tg_example}
The image below shows an example of what a snippet from the theme graph corresponding to the action class \textit{carving pumpkins}' would look like. Nodes represent tagged nouns, verbs, and adverbs from the ground-truth training annotations. All edges in the graph are undirected and weighted by normalised point mutual information scores.

\begin{figure}[H]
  \centering
  \includegraphics[width=1\linewidth]{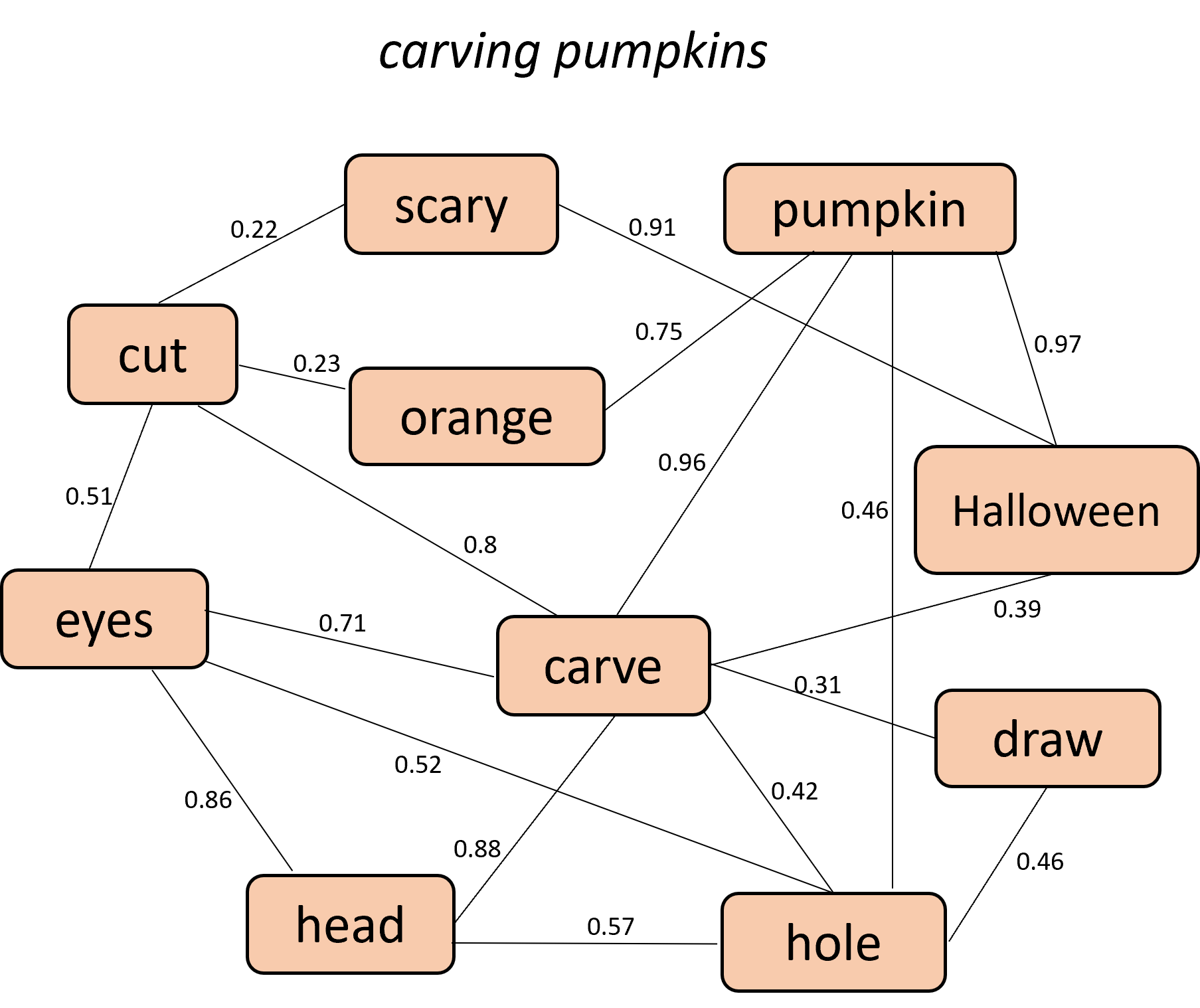}
   \caption{Visual example of a sub-graph for the theme graph corresponding to the ActivityNet action class \textit{carving pumpkins}.}
  \label{fig: tg_graph_example}
\end{figure}

\begin{table*}[t]
  \centering
  
  \resizebox{1\linewidth}{!}{
  \begin{tabular}{@{}l||c|l|l|c|cccc|c}
    \toprule
     \multicolumn{9}{c}{\textit{\textbf{ae-val}}}  \\ \hline
     \textbf{Model} & \textbf{Conference} & \textbf{Year} & \textbf{Modalities} & \textbf{Integration Method} & \textbf{B4} $\uparrow$ & \textbf{M} $\uparrow$ & \textbf{C} $\uparrow$  & \textbf{R} $\uparrow$ & \textbf{R4} \\ \midrule
     VTrans \cite{zhou2018end} & CVPR & 2018 & V+F & Concatenation & 9.75 & 15.64 & 22.16 & 28.9 &  7.79  \\
     HSE \cite{zhang2018cross} & ECCV & 2018 & V & - & 9.84 & 13.78 & 18.78 & - \\
     AdvInf \cite{park2019adversarial} & CVPR & 2019 & V+F+O & Concatenation & 10.04 & 15.93 & 27.27 & - &  5.76 \\
     GVD  \cite{zhou2019grounded} & CVPR & 2019 & V+F+O & CM Attention &  11.04 & 15.71 & 22.95 & - &  8.76 \\
     GVDsup \cite{zhou2019grounded} & CVPR & 2019 & V+F+O & CM Attention & 11.30 & 16.41 & 22.94 & - & 7.04 \\
     Trans-XL \cite{dai2019transformer} & ACL & 2019 & V+F & Concatenation & 10.39 & 15.09 & 21.67 & 30.18 &  8.79 \\
     Trans-XLRG \cite{lei2020mart} & ACL & 2020 & V+F  & Concatenation &  10.17 & 14.77 & 20.40 & - \\
     MDVC \cite{iashin2020multi} \dag & CVPR & 2020 & V+S+A & Concatenation & 9.12  & 14.69  &  17.57  & 25.85 & -  \\
     BMT \cite{iashin2020better} \dag & BMVC & 2020 & V+A & CM Attention & 9.00 & 14.49  & 16.46  & 26.11 & -  \\
     MART \cite{lei2020mart} &  ACL & 2020 & V+F & Concatenation & 10.33 & 15.68 & 23.42 & - &  5.18 \\
     PDVC \cite{wang2021end} & ICCV & 2021 & V+F & Concatenation & 11.8 & 15.93 & 27.27 & - &  - \\
     VLCAP \cite{yamazaki2022vlcap} & ICIP & 2022 & V+L & CM Attention & 14.00 & 17.78 & 32.58  & 36.37 & 4.42 \\
     Motion-Aware \cite{hu2023motion} & ICASSP & 2023 & V+O & CM Attention & 12.07 & 16.81 & 29.32 & - & \textbf{4.28} \\ 
     Text-KG \cite{gu2023text} & CVPR & 2023 & V+O+S+G(S+C) & CM Attention & 11.30  & 16.50 & 26.60 & - &  6.30 \\
     VLTinT w/ CL \cite{yamazaki2023vltint} & AAAI & 2023 & V+L+O & CM Attention & \textbf{14.93} & \textbf{18.16} & \textbf{33.07} & \textbf{36.86} & 4.87 \\ 
     VLTinT w/ CL$^\ast$ \cite{yamazaki2023vltint} & AAAI & 2023 & V+L+O & CM Attention & \underline{14.89} & \underline{18.09} & \textbf{33.07} & 
     \underline{36.76} & 5.11 \\ 
     VGCSN+CHPG \cite{yu2024exploring} & ICASSP & 2024 & V+L+O+C & CM Attention & 12.20 & 16.69 & 29.98 & - & \underline{4.32} \\ 
     \hline 
     GEM-VPC w/ No Recurrence & - & 2024 & V+G(V+A+C) & CM Attention & 13.16 & 17.56 & 27.50 &  33.85 & 7.86  \\
     GEM-VPC w/ MART decoder & - & 2024 &  V+G(V+A+C) & CM Attention & 13.91  & 17.47 & 30.68 & 35.97  &  6.18 
     \\
     GEM-VPC w/ TinT decoder & - & 2024 & V+G(V+A+C) & CM Attention & 14.73 & 18.02 & \underline{32.93} & 36.71 & 5.41
     \\
    \bottomrule
  \end{tabular}}
  \caption{$n$-gram metrics and repetition scores of baselines and our model (GEM-VPC) for the ActivityNet \textit{ae-val} split. In the `Modalities' column, the abbreviations are defined as follows: V=visual, F=optical flow, O=bounding box object visual features, A=audio, S=speech, L=language, G(V+A+C)=graph built with visual, audio modality and commonsense, G(S+C)=graph build with speech modality and commonsense. \dag \space indicates results computed by ourselves using VPC evaluation mode.$\ast$ indicates results computed from the model that was reran with our own environment. The `Integration Method' column indicates the model's main approach for integrating the distinct modalities. `Concatenation' refers to a simple concatenation of different modality vectors which are then fed into a single stream, `CM Attention' refers to cross-modal attention employed between modules processing different modality inputs, and `Joint CM Space' indicates that the model attempts to learn a common space for different modalities.}\label{tab: automatic_metrics_anet_valid}
\end{table*}

\section{Implementation Details} \label{implementation_details}
\textbf{Graph Construction:} The TimeSformer \cite{bertasius2021space} pretrained on the Kinetics600 dataset \cite{zisserman2017kinetics} was used as the action classification model for constructing the action nodes for the VF-method. The model is capable of predicting 600 unique action classes. We leveraged the Audio Spectrogram Transformer \cite{gong2021ast} pretrained on AudioSet \cite{gemmeke2017audio} (capable of predicting 632 audio event classes) as the audio classification model to create the audio nodes for the VF and ASR-method. Commonsense nodes are generated by Comet-ATOMIC2020 \cite{hwang2021comet} using the \textit{`comet\_atomic2020\_bart'} implementation. Object and location nodes for the VF-method are generated by the BLIP-VQA base model as proposed in \cite{li2022blip}, with the object nodes further expanded using Detic's \cite{zhou2022detecting} object detection model. ASR from the YouCook2 videos was extracted using OpenAI's Whisper \cite{radford2023robust} while we used AllenNLP's OpenIE model \cite{Stanovsky2018SupervisedOI} for creating the action nodes in the ASR-method. All part-of-speech tagging is done with the NLTK toolkit. 

For each set of commonsense knowledge generated by its corresponding action node, we filter out any similar generated commonsense to avoid adding duplicate commonsense into the video-specific graph at the same timestep. Specifically, we removed any similar commonsense if its Levenshtein Distance ratio with another commonsense is greater than 0.70. As mentioned in Section 3.1 of the main paper, we also did not add the commonsense into the graph if the action class used to generate that commonsense had a confidence score of less than 0.5 so as to avoid incorporating irrelevant external knowledge. The threshold for filtering out any noisy object and audio labels was 0.25 and 0.3 respectively while the threshold to determine whether an action node contained \textit{`no action'} was 0.35. For creating the theme graphs in the case when the ASR-method is used, $k$-means clustering with $k=300$ and 10 repetitions was used to create the action classes. The theme graphs contain the top-100 most occurring words within that action class/theme. 

\textbf{Model Training:} The 2048D visual features for the ActivityNet were extracted using a 3D-CNN backbone \cite{ji20123d}. For YouCook2, we used 2048D ResNet-200 \cite{he2016deep} visual features concatenated with 1024D optical flow features from BNInception \cite{ioffe2015batch}.  The node/edge linguistic features for the video-specific and theme graphs are represented using CLIP textual embeddings \cite{radford2021learning}.

We train the modules in an end-to-end fashion with teacher forcing to optimise the Kullback–Leibler divergence loss with the best model using a label smoothing of 0.3. The word embedding matrix of the models is initialised with GloVe embeddings of dimension 300 \cite{pennington2014glove}. Inputs into each transformer stream are added with fixed positional embeddings (only for the visual stream) and learnt token type embeddings. The token type embedding matrix was size 10 to incorporate for different modality types such as visual, audio or type of node e.g. location, commonsense etc. We use 2 hidden transformer layers with 12 attention heads where the hidden and intermediate size was 768. For the theme graph encoder, 2 GATv2Conv layers \cite{brody2021attentive} were used while the video-specific graph encoder used 1 GATv2Conv layer with all layers using 4 attention heads. Adam optimizer was used to train our model with an initial learning rate of 1e-4, $\beta_1 = 0.9$ and $\beta_2$ = 0.999, $L_2$ weight decay of 0.01, learning rate warmup over the first 5 epochs and batch size of 2. Early stopping was applied after no improvement was seen in the validation CIDEr score in 3 consecutive epochs. For decoding the caption at inference, nucleus sampling with 0.6 top-$p$ and 0.5 temperature was used. 

\section{ActivityNet Validation Set Quantitative Results} \label{ae-val scores}
Table \ref{tab: automatic_metrics_anet_valid} shows the $n$-gram metrics and repetition scores of baselines and GEM-VPC for the ActivityNet \textit{ae-val} split. In the `Modalities' column, the abbreviations are defined as follows: V=visual, F=optical flow, O=bounding box object visual features, A=audio, S=speech, L=language, G(V+A+C)=graph built with visual, audio modality and commonsense, G(S+C)=graph build with speech modality and commonsense. \dag \space indicates results computed by ourselves using VPC evaluation mode.$\ast$ indicates results computed from the model that was reran with the same environment as this research. The `Integration Method' column indicates the model's main approach for integrating the distinct modalities. `Concatenation' refers to a simple concatenation of different modality vectors which are then fed into a single stream, `CM Attention' refers to cross-modal attention employed between modules processing different modality inputs, and `Joint CM Space' indicates that the model attempts to learn a common space for different modalities.

Our best model (GEM-VPC w/ TinT decoder) achieves comparable performance with the strongest baselines (VLTinT w/ CL and VLTinT w/ CL$^\ast$). Note that while we underperform slightly on the validation set, we outperform VLTinT in a majority of the metrics when evaluating on the testing set (see Table 1 of the main paper). 
\\
\\
\textbf{Please note that the appendix continues on the next page.}

\onecolumn
\newpage
\section{Video-Specific Graph Visual Examples} \label{appendix: vg_visual_examples}
Visual depiction of what the video-specific graphs would look like using the VF and ASR-method for an example ActivityNet and YouCook2 video. Blue nodes represent the action nodes, red nodes are the location/contextual phrase nodes, green nodes are object nodes, purple nodes are audio nodes and orange nodes are the commonsense nodes. Note that due to size of the graphs, not all nodes are presented and graphs would be larger in reality. Sentences under the video frames are the matching ground-truth captions. 
\raggedbottom
\begin{figure}[H]
  \centering
  \includegraphics[width=1\linewidth]{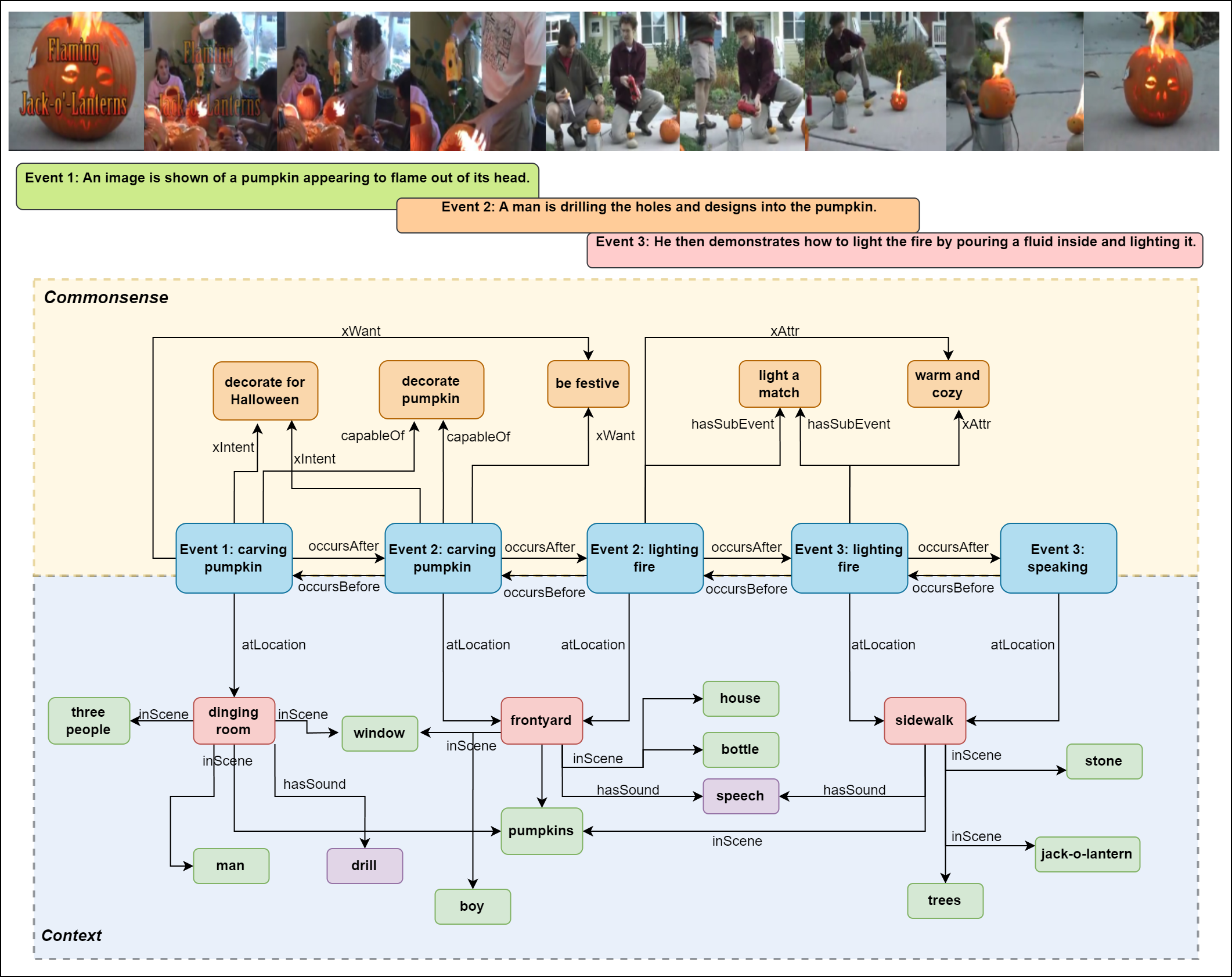}
  \caption{Video-specific graph for an example video in the ActivityNet dataset using the VF-method for the first 3 timesteps.}
  \label{fig: vg_graph_carving_pumpkins}
\end{figure}

\begin{figure}[H]
  \centering
  \includegraphics[width=1\linewidth]{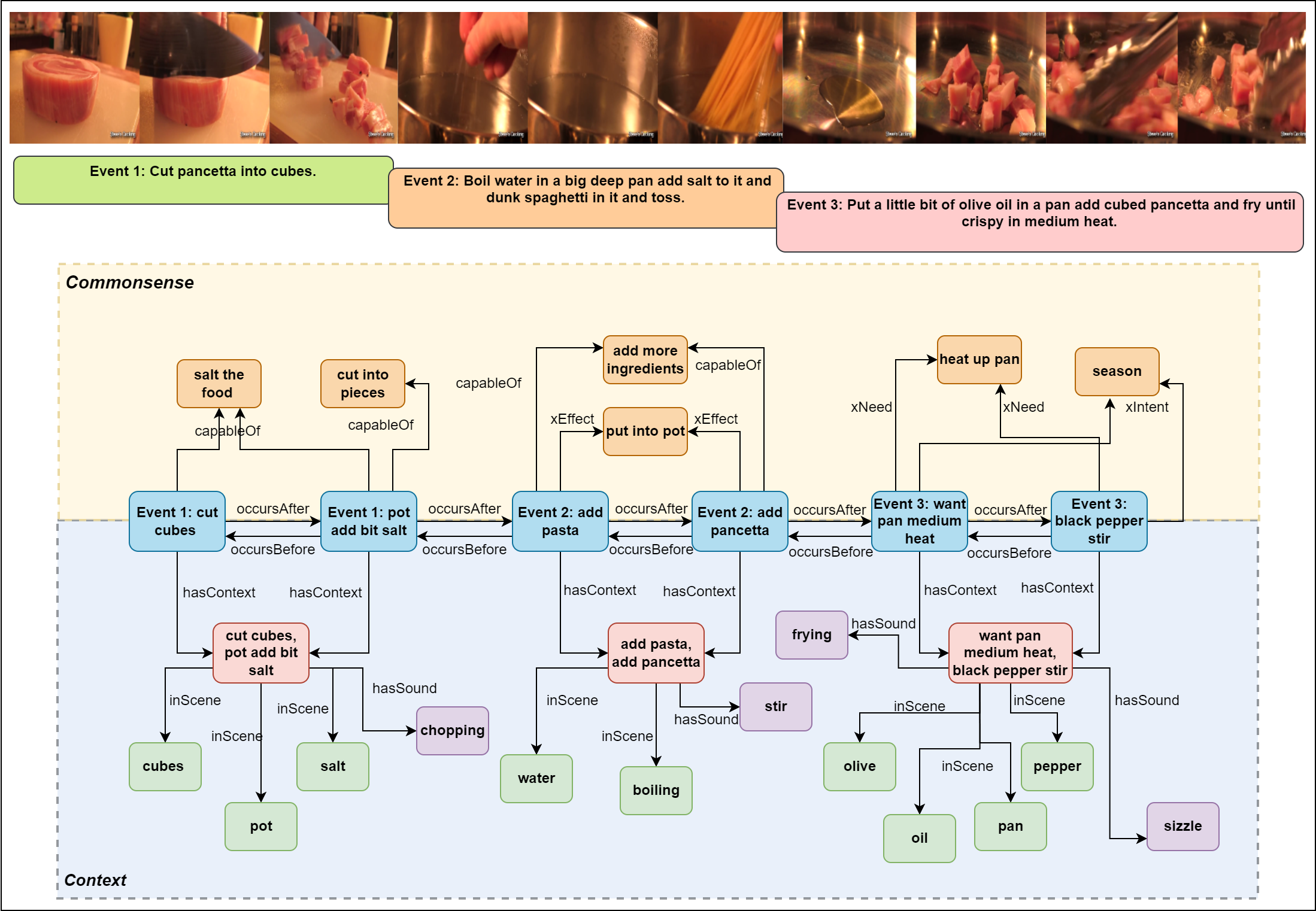}
  \caption{Video-specific graph for an example video in the YouCook2 dataset using the ASR-method for the first 3 timesteps.}
  \label{fig: vg_graph_making_pasta}
\end{figure}

\newpage 
\section{Qualitative Examples (Ours vs SOTA)} \label{appendix: qual_examples}
\begin{figure}[H]
  \centering
  \includegraphics[width=1\linewidth]{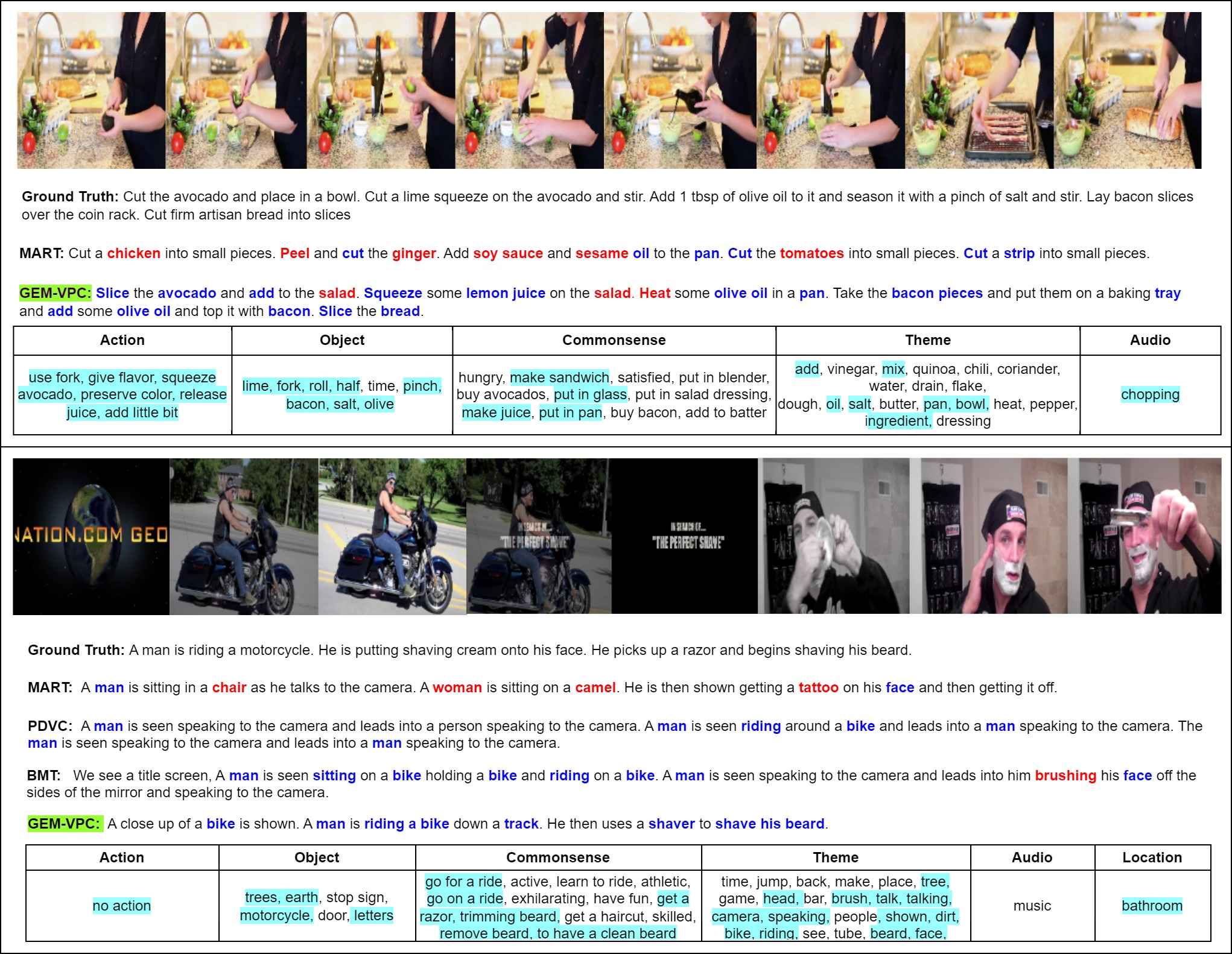}
  \caption{Qualitative Examples for the state-of-the-art methods versus ours.}
  \label{fig: qualitative_examples}
\end{figure}

Qualitative Examples for the start-of-the art methods versus ours (GEM-VPC) are shown in Figure \ref{fig: qualitative_examples}. The first example is from YouCook2 while the last is from ActivityNet. Blue words in the machine-generated captions are visually grounding to the video, while red words represent irrelevant words that are `hallucinated' by the model. We collect the top-10 selected nodes by confidence score at each timestep during inference and display the selected nodes and their types in the table after each example. Highlighted blue words in the table indicate information related to the theme of the video.

\section{More Qualitative Examples}
\begin{figure}[H]
  \centering
  \includegraphics[width=1\linewidth]{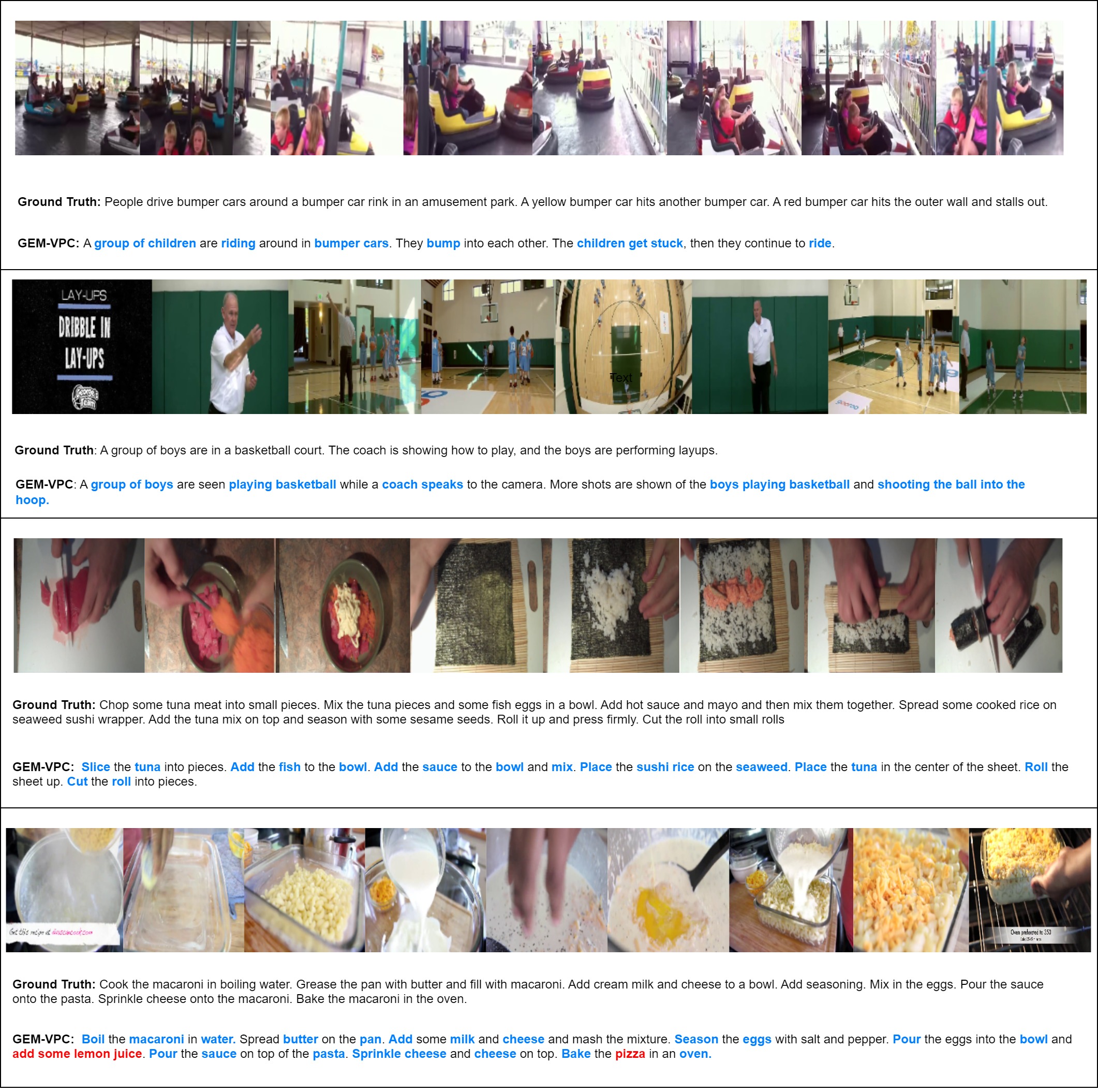}
  \caption{Qualitative examples of generated captions using our model. Top 2 examples are from ActivityNet and bottom 2 examples are from YouCook2. Blue words in the machine-generated captions are visually grounding to the video, while red words represent irrelevant words that are `hallucinated' by the model.}
  \label{fig: qualitative_examples2}
\end{figure}
\end{document}